\documentclass[runningheads]{llncs}
\usepackage{graphicx}

\usepackage{tikz}
\usepackage{comment}
\usepackage{amsmath,amssymb} 
\usepackage{color}
\usepackage{stmaryrd}
\usepackage{cite}
\usepackage{subfig}
\usepackage{hyperref}
\hypersetup{
    colorlinks=true,
    linkcolor=blue
}
\usepackage{multirow}
\usepackage{booktabs}
\usepackage[accsupp]{axessibility}  

\usepackage{dsfont}

\newif\ifshowedits

\newcommand{\addeditor}[3]{%
  \definecolor{#1color}{rgb}{#3}
  \expandafter\newcommand\csname #1\endcsname[1]{
  \ifshowedits
    {\color{#1color} ##1}
  \else
    {##1}
  \fi
  }%
  \expandafter\newcommand\csname #1rmk\endcsname[1]{
  \ifshowedits
    {\color{#1color} {\bf [#2: ##1]}}
  \fi
  }%
}

\newcommand{\calH}{{\cal H}}

\newcommand{\calL}{{\cal L}}

\DeclareFontFamily{U}{mathx}{\hyphenchar\font45}
\DeclareFontShape{U}{mathx}{m}{n}{
      <5> <6> <7> <8> <9> <10>
      <10.95> <12> <14.4> <17.28> <20.74> <24.88>
      mathx10
      }{}
\DeclareSymbolFont{mathx}{U}{mathx}{m}{n}
\DeclareFontSubstitution{U}{mathx}{m}{n}
\DeclareMathAccent{\widebar}{0}{mathx}{"73}

\addeditor{vincent}{VL}{0.0, 0.5, 0.0}
\addeditor{philippe}{PC}{0.0, 0.0, 0.5}
\addeditor{eperot}{EP}{1.0, 0.3, 1.0}
\addeditor{amos}{AS}{1.0, 0.6, 0.0}
\showeditstrue

\begin{document}
\pagestyle{headings}
\mainmatter

\title{Long-Lived Accurate Keypoint\\in Event Streams}

\titlerunning{Long-Lived Accurate Keypoints\\in Event Streams}
%
\author{Philippe Chiberre\inst{1,2} \and
Etienne Perot \inst{1} \and
Amos Sironi \inst{1} \and
Vincent Lepetit \inst{2}}
\authorrunning{Philippe Chiberre, Etienne Perot, Amos Sironi and Vincent Lepetit}
%
\institute{Prophesee, Paris, France \and
Ecole des Ponts, Univ Gustave Eiffel, CNRS, LIGM, F-77454 Marne-la-Vallée, France
\email{\{pchiberre,asironi\}@prophesee.ai}
\email{vincent.lepetit@enpc.fr}}

\maketitle

\begin{abstract}

We present a novel end-to-end approach to keypoint detection and tracking in an event stream that provides better precision and much longer keypoint tracks than previous methods. 
This is made possible by two contributions working together. 
First, we propose a simple procedure to generate stable keypoint labels, which we use to train a recurrent architecture. This training data results in detections that are very consistent over time. 
Moreover, we observe that previous methods for keypoint detection work 
on a representation (such as the time surface) that integrates events over a period of time. Since this integration is required, we claim it is better to predict the keypoints' \emph{trajectories} for the time period rather than single locations, as done in previous approaches. We predict these trajectories in the form of a series of heatmaps for the integration time period. This improves the keypoint localization. 
Our architecture can also be kept very simple, which results in very fast inference times.  We demonstrate our approach on the HVGA ATIS Corner dataset as well as ``The Event-Camera Dataset and Simulator'' dataset, and show it results in keypoint tracks that are three times longer and nearly twice as accurate as the best previous state-of-the-art methods.  We believe our approach can be generalized to other event-based camera problems, and we release our source code to encourage other authors to explore it.


%
\end{abstract}



\section{Introduction}

\begin{figure}
    \centering
    \subfloat[\centering]{{\includegraphics[width=5.6cm]{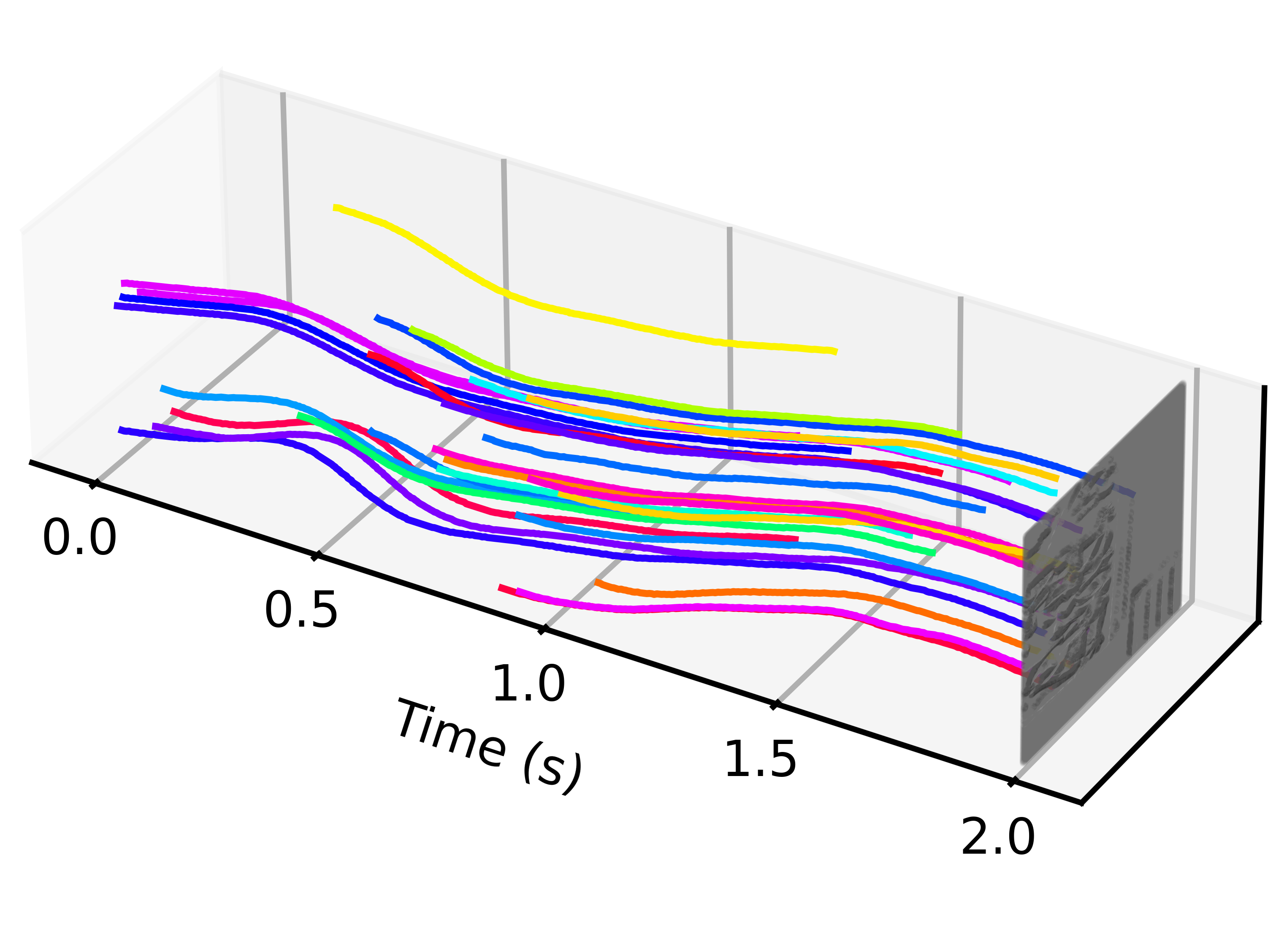} }}%
    \qquad
    \subfloat[\centering]{{\includegraphics[width=5.6cm]{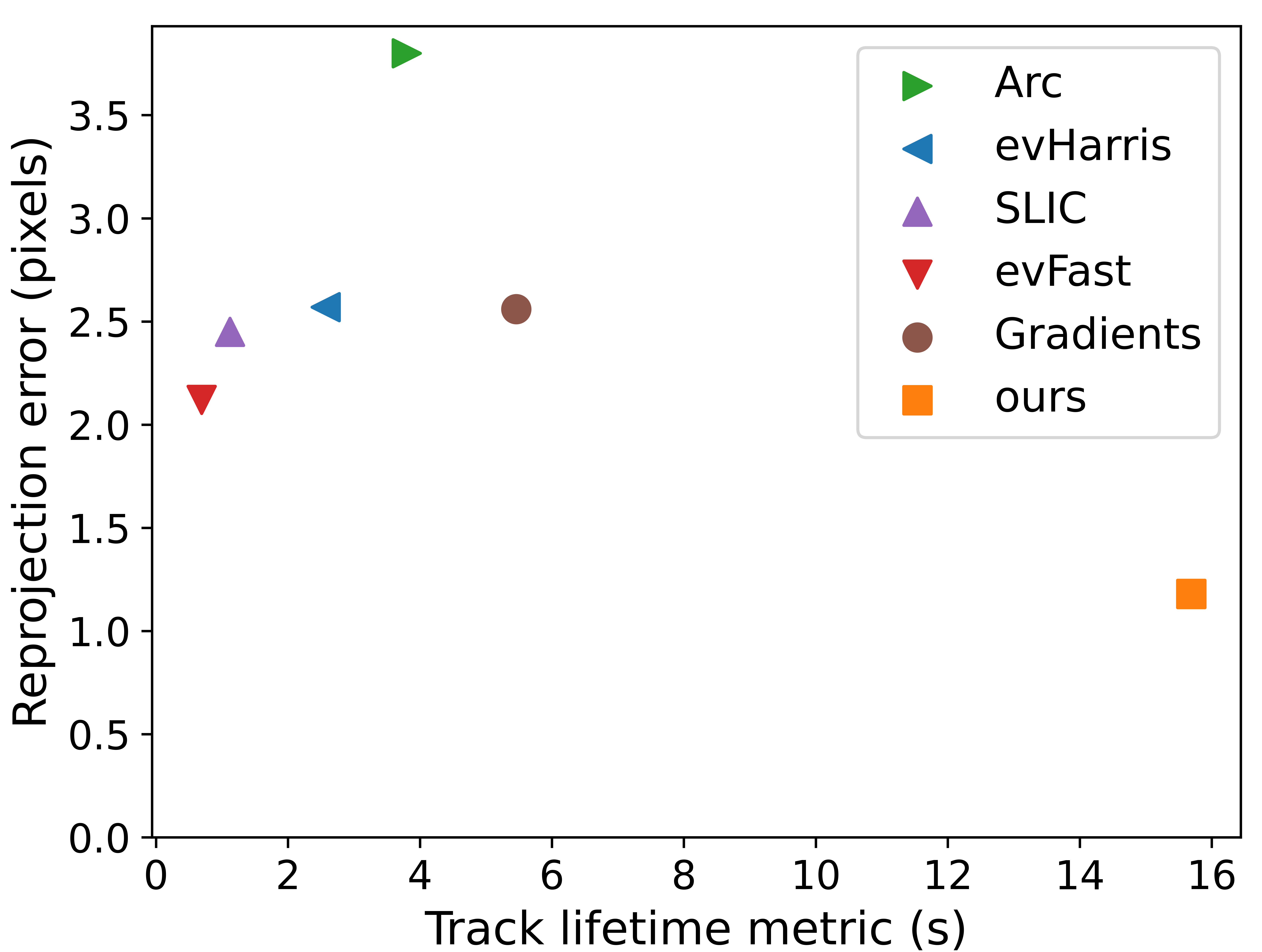} }}%
    \caption{
      (a) Our tracks for the beginning of the Guernica sequence from the HVGA ATIS Corner dataset~\cite{manderscheid2019speed}.
      (b) We compare ourselves to Arc~\cite{arc}, evHarris~\cite{eharris}, SILC~\cite{manderscheid2019speed}, evFast~\cite{efast}, and the Gradients-based method of \cite{gradients}. Our detector predicts well-localized keypoints that can be tracked reliably with a
      simple nearest-neighbor matching algorithm to  obtain very long tracks. It
      nearly triples the track lifetime metric of  the previous best method  while reducing the reprojection error by more than 1 pixel.
      }
    \label{fig:main_vis}
\end{figure}


Keypoint detection is a fundamental problem of computer vision and a building block to a large number of applications such as Simultaneous Localisation and Mapping~(SLAM), Structure-from-Motion~(SfM), object recognition and tracking. Event-based cameras are recent sensors capturing the change in luminosity, asynchronously across the sensor~\cite{Lichtsteiner2008,son20174,finateu20205}.  They are able to capture high dynamic range scenes while maintaining a low-power consumption, making them particularly attractive for embedded systems~\cite{Gallego2020}. Not surprisingly, many methods for keypoint detection for event-based cameras have already been proposed~\cite{efast, eharris, arc, manderscheid2019speed, gradients, luvharris}.

However, the nature of the signal provided by event-based cameras is very different from the images captured by regular cameras.  Event-based cameras send 'events', which signal a sudden change of light intensity at an image location. Alone, an event therefore provides very little information.  Events are also very noisy, with many false positives and false negatives~\cite{Delbruck2020-sb,Graca2021}.  This makes computer vision problems for event-based cameras particularly challenging.

As shown in Figure~\ref{fig:main_vis}, we propose a novel event-based keypoint detector that significantly outperforms state-of-the-art methods, both in terms of stability as it provides much longer tracks, and accuracy as the keypoints are much better localized.

Our detector is based on a deep recurrent architecture. A first simple but critical contribution is how we generate training data: We generate video sequences of bitmap frames by applying homographies to images from the COCO dataset~\cite{coco}. We transform these video sequences into event streams using an event-based camera simulator. To obtain keypoint location labels, we run the Harris corner detector~\cite{Harris88} on the original image from COCO. We warp these locations using the homographies to obtain the keypoint locations over time. This generates much more stable labels than, for example, running the Harris detector on all the frames of the sequences. 

While this procedure is simple, this results in  keypoint detections that are much more consistent over time than previous methods. These detections can be linked by a simple nearest-neighbor matching procedure into keypoint tracks very reliably. This results in very long keypoint tracks, which are very important for many applications such as object tracking or Structure-from-Motion. This also provides very accurate locations. Note that while our architecture is trained on planar scenes (since we use homographies), it also works well on 3D scenes, which we show by evaluating on the dataset from The Event-Camera Dataset and Simulator~\cite{zurich_event_dataset}.  Generalization of keypoints trained on planar scenes to 3D scenes was observed before for bitmap frames, for example in \cite{superpoint}.

We also observed all previous event-based keypoint detection methods integrate events over a period of time in a 2D buffer. This integration period is required to gather enough information from the events, and compensates for noise. While this integration period is required, all previous methods for keypoint detection for event-based cameras still provide a single image location for each keypoint detected for this period.  This is probably only because of the legacy of keypoint detection methods for bitmap images, in which it is assumed that the image information is captured instantly. 

\begin{figure*}[t]
\centering
\begin{tabular}{ccccc}
    \includegraphics[width=.19\linewidth]{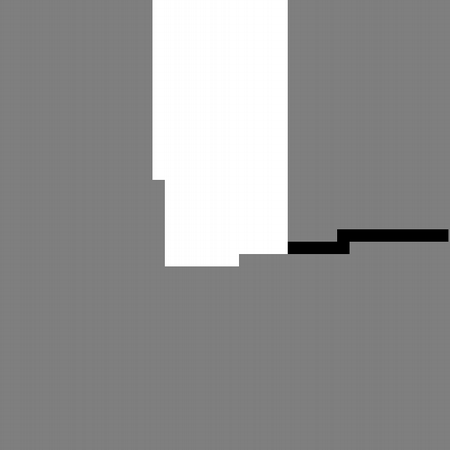}&
    \includegraphics[width=.19\linewidth]{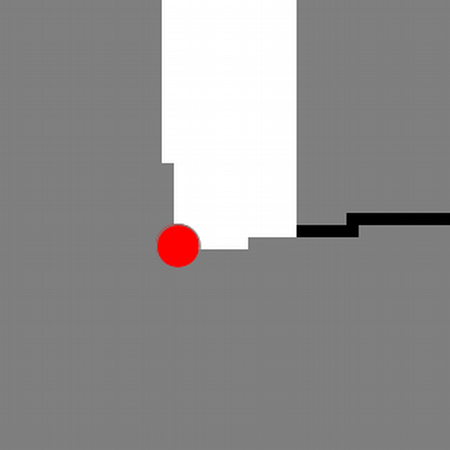}&
    \includegraphics[width=.19\linewidth]{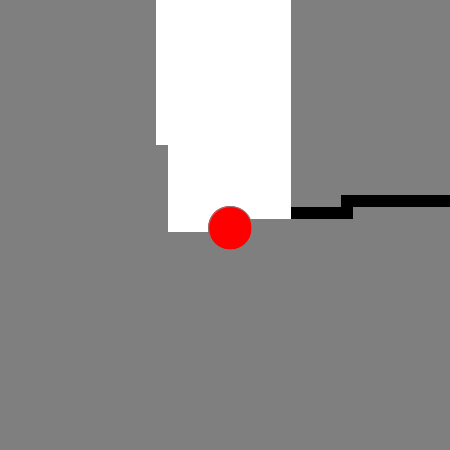}&
    \includegraphics[width=.19\linewidth]{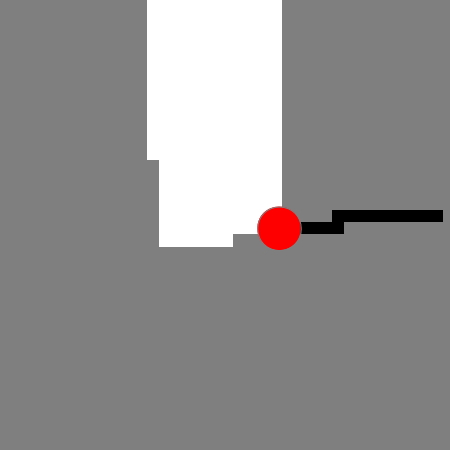}&
    \includegraphics[width=.19\linewidth]{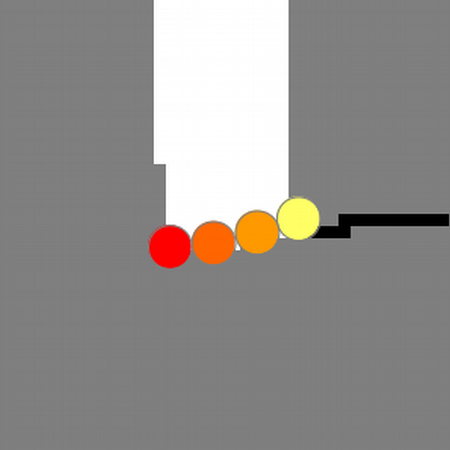}\\
    (a) & (b) & (c) & (d) & (e) \\
\end{tabular}    
    \caption{\label{fig:insight} {\bf Keypoint detection as trajectory prediction. } 
    (a) Buffer of accumulated events around a keypoint over a short period of time~(20ms). Note that the keypoint is actually spread over several image locations. (b-d) Some of the possible locations for the keypoint, if one sticks to a single location for the buffer. (e) Instead, we predict multiple heatmaps each corresponding to a point in time (shown here with a gradient in color). This makes linking detections from several time periods much more reliable, and results in much longer keypoint tracks.
    }
\end{figure*}

Instead, we predict multiple successive locations rather for a single integration period. This is illustrated in Figure~\ref{fig:insight}: We predict a series of successive heatmaps for the entire frame. The local maxima of the heatmaps gives us the predicted keypoint locations. Predicting multiple heatmaps lets us handle different numbers of keypoints and the fact that keypoints can appear or disappear during the integration period. Because the predicted locations are spaced out by a very fine time step, we significantly improve the accuracy of our predictions. 

An alternative would be to predict a single location per keypoint, but for many overlapping integration periods. However, this would result in significantly increased computation times. By contrast, our approach is very light and has a computational cost similar to previous methods. 


Beyond keypoint detection, we believe that our first observation---predicting multiple successive estimates for the integration period rather than a single one---can be applied to other event-based cameras such as segment detection, depth prediction and object detection to improve their accuracy. Thus, we hope our work will encourage other researchers to explore this direction.

\section{Related Work}

We focus here on  methods for keypoint detection in event  streams.  They can be classified  into two  categories: The  first category  is made  of 'handcrafted' methods, in contrast to methods based on machine learning.







\subsection{Handcrafted Methods}

Multiple hand crafted methods for event-based are inspired or adapted from frame-based feature detection. The easiest approach is to create a 2D intermediate representation on which one can apply a feature detector. eHarris~\cite{eharris} creates a binary image from the latest events, with pixel locations set to 1 if an event has occurred  during some integration period and set to 0 elsewhere. The Harris corner detector, originally defined for greylevel images~\cite{Harris88},  is then applied to this binary image. A modification of this approach has recently been proposed by Glover et al. with luvHarris~\cite{luvharris}. They introduce a novel representation of events, named Threshold-ordinal Surface, replacing the binary image. This new representation makes the Harris detection more precise and more stable at the expense of a greater computational load: the representation is computed event-by-event, which is not parallelisable. The computational impact is lessened by only rendering the Harris score map "as-fast-as-possible" and achieving real-time detection. 

eFAST~\cite{efast} and ARC~\cite{arc} on the other hand use a non-computationally intensive representation, named the Surface of Active Events (SAE). The SAE is defined for each pixel and polarity as the latest time an event has occurred:  
\begin{equation}
\text{SAE}[x,y,t] \leftarrow t \> .
\end{equation}
The SAE is also computed for an integration period since it contains the information of many preceding events.
They then proceed to apply a local matching pattern for each event at its location in the SAE. \cite{efast} considers two circles around the latest event and compares their timestamps. In \cite{arc}, Alzugaray and Chli also consider two concentric circles around the latest event but introduce new rules for the classification of an event as a corner. \cite{arc} achieves better results while at the same time requiring less computations. In all cases, the approach of computation event-by-event becomes quickly intractable as the computation cannot be done in parallel while at the same time, resolution of event-based sensors keep increasing \cite{finateu20205} and event rates can increase dramatically.

Hand-crafted approaches are sensitive to the noise present in the event stream. This is why recently, several authors modeled the keypoint detection problem as a machine learning one.

\subsection{Learned Methods}

\cite{manderscheid2019speed} introduced a novel event representation, called the Speed-Invariant Time Surface. It aims at removing the impact of motion speed on the differences of appearance of the same keypoint in an event stream. Instead of applying a frame-based feature detector, they train a random forest to classify each event as a keypoint or a non-keypoint. Despite improving the quality of the keypoints, the computation of the representation and random forest make this approach unsuitable in practice.

In~\cite{e2vid}, the authors show that a gray-level image can be reconstructed from events by training a recurrent neural network. Standard frame-based keypoint detectors can be applied on the output of their method, leading to very accurate results. More recently, \cite{gradients} observed that reconstructing a frame is unnecessary and adds computational complexity. In fact, the authors noticed that events are closely related to image gradients, and proposed to learn to predict them and use them to compute the Harris score. One advantage of this work is that the deep architecture used to predict the gradients is very light, and we use a similar architecture to predict the heatmaps.

\vspace{0.5cm}

As discussed in the introduction, all these works rely on a period of integration of events to output keypoint locations, but predict a single location per keypoint for this period. In this work, we propose to predict multiple successive locations, and show this leads to much more accurate keypoint localization.


\section{Method}

\begin{figure*}[t]
    \begin{center}
    \includegraphics[width=.99\linewidth]{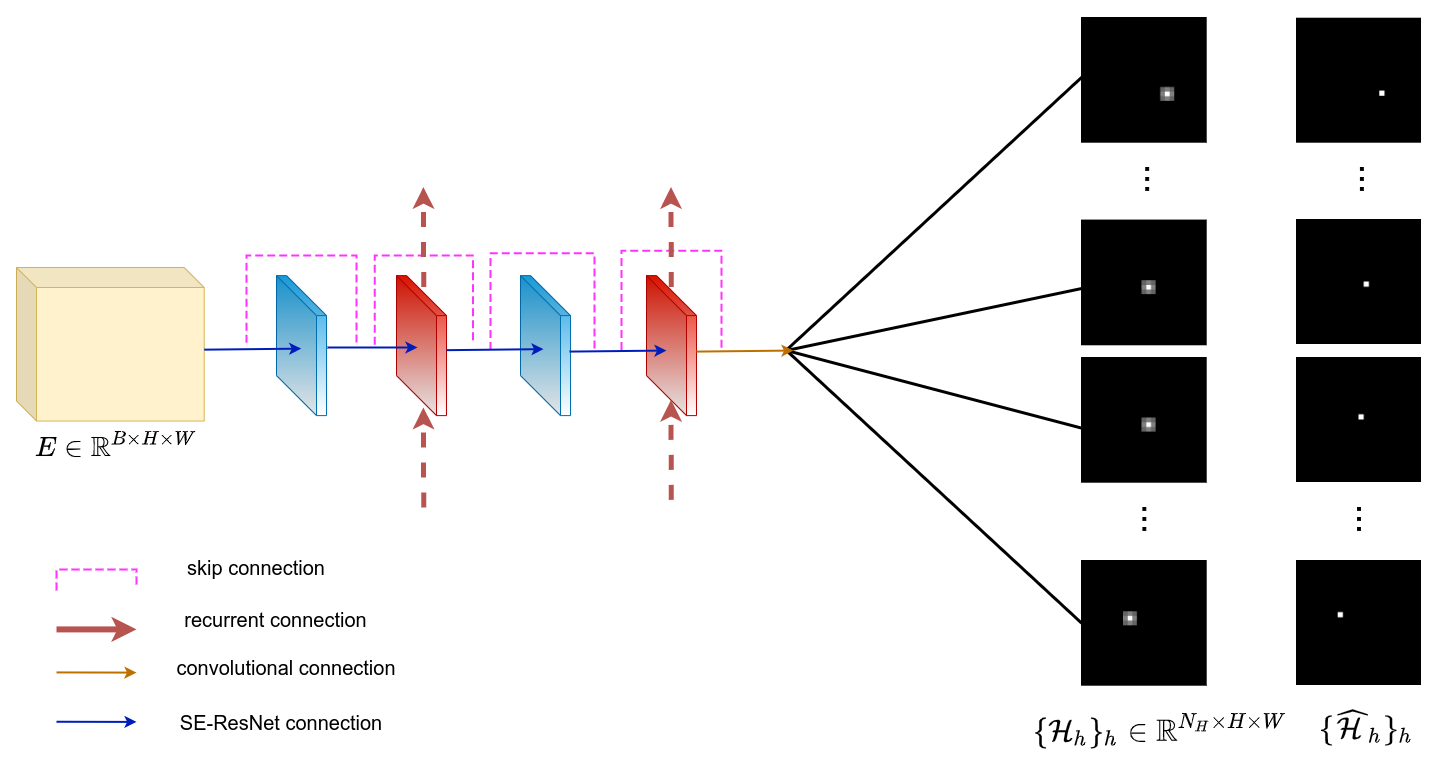}
    \end{center}
    \caption{{\bf Overview of our inference pipeline.}  An event cube is build from the input events and fed to a recurrent neural network. We keep the architecture simple and efficient. We use the one introduced in  \cite{firenet} with the same modifications as in \cite{gradients}. 
    More precisely, the event cube is given as input to a Squeeze-and-Excite ResNet block, followed by a ConvLSTM with residual connection. This block of two layers is repeated once. Finally, a simple convolutional layer predicts the keypoints trajectory as a sequence of $N_{H}$ heatmaps.
    }
    \label{fig:architecture}
\end{figure*}

Our detector integrates the events sent by the camera over a time period into an `event cube', and outputs a series of heatmaps for this time period, Figure~\ref{fig:architecture}. We detail below how we construct this event cube in practice, the nature of the heatmaps, the deep architecture we use, and how we generate training data to train it.

\subsection{Input Event Representation}
\label{event_cube}

The input to our detector is a $H \times W \times B$ event tensor $E(x, y, t)$, where  $H$, $W$ are the image sensor height and width respectively and $B$ is the number of temporal bins. In practice, we use $B = 10$. We use the method first proposed in \cite{timebins} and describe it briefly below for completeness.

Each input event $(x_i, y_i, t_i, p_i)$ received during the integration period $\Delta T$ contributes to $E$ by its polarity $p_i$ to the two closest temporal bins using a triangular kernel. Formally, $E$ is computed as
\begin{equation}
    E(x, y, t) = \sum_i p_i \max(0, 1 - |t - t_i^\star|) \> ,
\end{equation}
for all image locations $x,y$ and all temporal bins $t$. The sum is over all the events such that $x_i = x$ and $y_i = y$. $t_i^\star$ is the normalized timestamp of the $i^{th}$ event:
\begin{equation}
    t_i^\star = \frac{(t_i - t_\text{min})}{\Delta T} (B - 1) \> ,
\end{equation}
where $t_\text{min}$ is the time at the beginning of the integration period.


\subsection{Predicting Heatmaps}

From the event cube, our detector predicts a set of heatmaps.
Heatmaps are convenient to predict the keypoints' locations as their number varies over time. In addition, predicting a dense tensor from a dense input like the event cube is standard and straightforward. 

The local maxima of the heatmaps are expected to correspond the keypoints' locations. To do so, we rely on the cross-entropy between the predicted heatmaps and the labelled locations for the keypoints:
\begin{equation}
\calL = \sum_{h\in[1;N_{H}]} \sum_{(x,y)} \text{BCE}(\calH_h(x,y), \widehat{\calH}_h(x,y)) \>.
\end{equation}
The first sum is over the $N_{H}$ predicted heatmaps, in practice we use $N_{H}=10$. The second sum is over the image locations $(x,y)$. $\text{BCE}$ denotes the binary cross-entropy. 




The $\widehat{\calH}_h$'s are labelled binary heatmaps for a given event cube: A location in $\widehat{\calH}_h$ is set to 1 if there is a keypoint at this location at time step 
$[t_\text{min} + (h-1)/N_{H} \times \Delta T, t_\text{min} + (h)/N_{H} \times \Delta T[$
and set to 0 otherwise. The $\calH_h$'s are the predicted heatmaps for the event cube. We guarantee that their values remain between 0 and 1 by applying the logistic function on the last layer of our architecture. A similar loss function was used in \cite{s2dnet} for example. It encourages the local maxima with large values in the predicted heatmaps to correspond to the locations of the keypoints.

\begin{figure*}[t]
    \begin{center}
    \includegraphics[width=.99\linewidth]{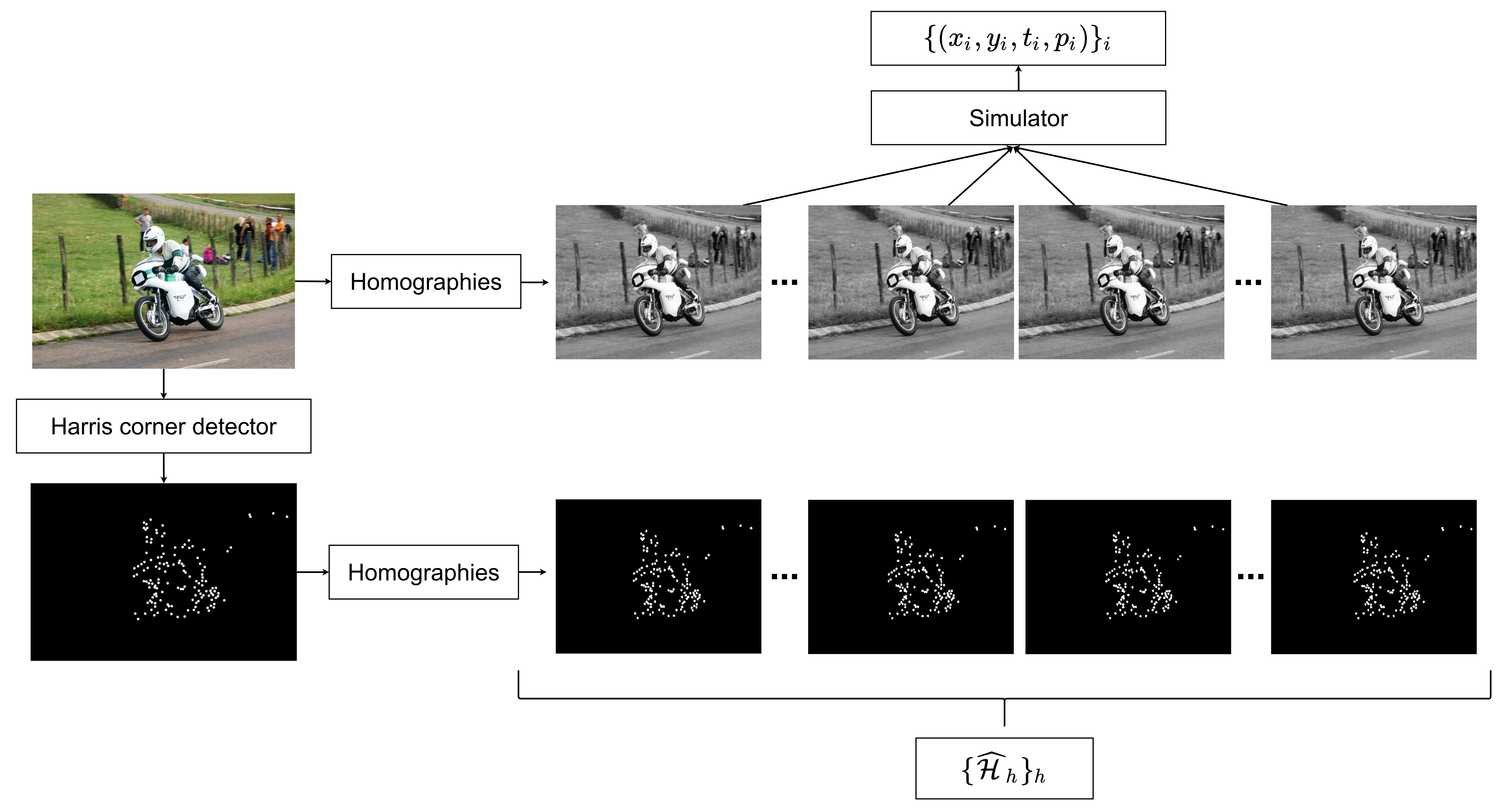}
    \end{center}
    \caption{{\bf Generating training data.} Similarly to~\cite{gradients}, given a still image, we apply homographies to warp it and generate a video sequence. However, by contrast with~\cite{gradients}, we use the same homographies to warp the heatmap of Harris' corner of the still image to have a perfect match between the video frames and the corners. From the video sequence we then generate events using a simulator and then compute the event cubes~\cite{timebins} used as input to our network.
    }
    \label{fig:training_set}
\end{figure*}

\subsection{Creating Training Data}
\label{subsec:training_data}

To generate training data, we generate synthetic videos by applying homographies to grayscale images, and convert them into event streams using a simulator, as detailed below. The keypoint labels are obtained by applying the Harris corner detector to the original grayscale images. This procedure is also illustrated in Figure~\ref{fig:training_set}.

\paragraph{From an image to a video.} More precisely, to generate a synthetic video, we select an image from the COCO dataset~\cite{coco} at random. We apply the Harris corner detector to this image to obtain a set of keypoints. Then, as shown in Figure~\ref{fig:training_set}, we create a smooth video from homographies varying randomly to simulate complex camera motions in front of a planar scene. 


To do so, we first generate a number of sine waves with differing periods and phases. We use this signal as a basis for a random yet continuous translation vector and a rotation vector. By combining the translation and rotation vectors with a random depth, we can obtain a homography matrix. By applying these homographies to the image, we obtain a smooth yet random video. More details are given in the supplementary material.

We generate a homography and thus a video frame every $\Delta T / N_{H}$ seconds, where $\Delta T$ is the integration period, and $N_{H}$ the number of predicted heatmaps for this period. 

\paragraph{From the video to the event stream.} 
To generate events from the synthetic video, we apply our home-brewed simulator, which is mostly  a combination of \cite{rebecq2018esim} and \cite{Delbruck2020-sb}. 
This simulator computes the difference between consecutive frames after applying a logarithm to the pixel intensities.
It then generates events when the ON-or-OFF threshold is crossed by the log-difference, if the distance in time since the last event is greater than a hyperparameter for the refractory period. Multiple noise patterns are also added including random events firing, events firing multiple times and certain events always ON or OFF. The internal parameters of the simulator which define the noise are randomly selected for each video. 


This finally gives us a stream of events $\{(x_i, y_i, t_i, p_i)\}_i$, from which we can build the event cubes.

\paragraph{Generating the keypoint labels.}

We simply apply the homographies already used to generate the video frames to the keypoint locations in the COCO images.
From this, we can generate the $\widehat{\calH}_h$ heatmaps needed to compute the loss function. The ablation study in Table \ref{table:detect_vs_warp} shows the beneficial influence of warping the keypoints locations as opposed to detecting them independently.


\subsection{Training Details}
\label{training details}


We notice it is important to use hard negative mining during training. This is not surprising because of the imbalance between keypoint and non-keypoint pixels: Without mining, the network converges to the trivial solution of never predicting any keypoint, as this corresponds to a low value for the loss already.

At each iteration of the optimization, we select a labelled heatmap $\widehat{\calH}_h$, and build a temporally consistent batch that mixes keypoint and hard non-keypoint pixels. This batch is made of all the positive locations in $\widehat{\calH}_h$, \textit{i.e.} the $(x,y)$ such that $\widehat{\calH}_h(x,y)=1$ and the negative locations $(x,y)$ for which the current predicted value $\calH_h(x,y)$ is particularly wrong. More exactly, we use negative locations such that $\calH_h(x,y) > \tau$ with threshold $\tau$ selected such that the number of negative locations in the batch is three times the number of positive locations.




We also use truncated-backpropagation-through-time of 10 time steps,  detaching the state at each batch and never resetting for each video. We train for 30 epochs with a learning rate of $1e^{-4}$.

\subsection{Inference and Tracking}

At inference, we create the event cubes from incoming events over time periods. To define the time period, we use either a fixed length, or a fixed number of events. We apply our network to each event cube once it is built. We obtain $N_{H}$ heatmaps $\calH_h$, to which we apply a local non-maximum suppression for every patch of $7\times7$ pixels and keep every local maximum over a threshold $\tau_{keypoint}=0.2$ to be classified as a keypoint. 

To obtain keypoint tracks, we rely on a simple nearest neighbor tracking algorithm: For each heatmap successively and for each keypoint in this heatmap, we look for a neighbor in 
a small region (we consider a $9\times9$ region) and a short time period in the past (we consider a 7ms period). If a neighbor keypoint is found, we associate the new keypoint to its track. If more than one neighbor keypoint are found, we consider the closest one only. If we do not find any neighbor, we start a new track with the new keypoint.

\subsection{Architecture}

We use a very simple recurrent neural network to predict heatmaps $\{\calH_h\}_h$ from an event cube. Thanks to this simplicity, inference is very fast. We use an architecture similar to the one of \cite{gradients} except we predict multiple heatmaps instead of two gradient maps as done in \cite{gradients}.  It is shown in Figure~\ref{fig:architecture}, and we describe it here for completeness. Also note that \cite{gradients} still needs to compute the Harris score from the predicted gradients, while we directly predict keypoint heatmaps as output of the network.

Our architecture is a 5-layer fully convolutional network of $3\times 3$ kernels. Each layer has 12 channels and residual connections~\cite{resnet}. The second and fourth layers are  ConvLSTMs. The last layer, which predicts the heatmaps, is a standard convolutional layer, while the remaining feed-forward ones are Squeeze-Excite~(SE) connections~\cite{squeeze}.

\begin{figure*}[t]
\centering
\begin{tabular}{c@{ }c@{ }c@{ }c}
    \includegraphics[width=.24\linewidth]{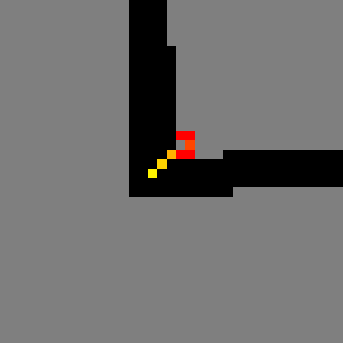}&
    \includegraphics[width=.24\linewidth]{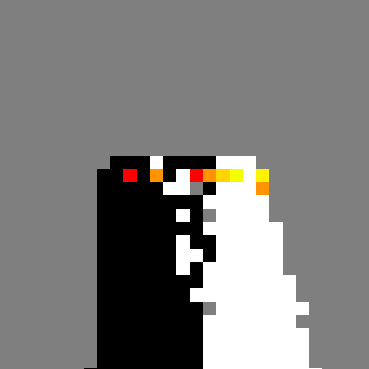}&
    \includegraphics[width=.24\linewidth]{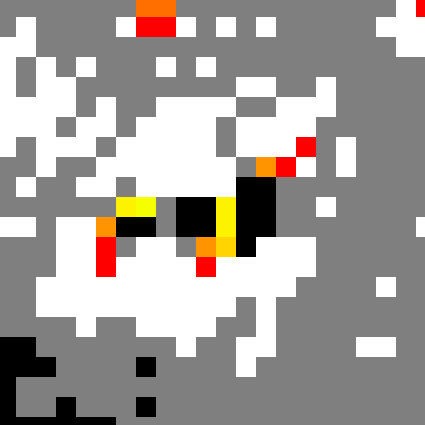}&
    \includegraphics[width=.24\linewidth]{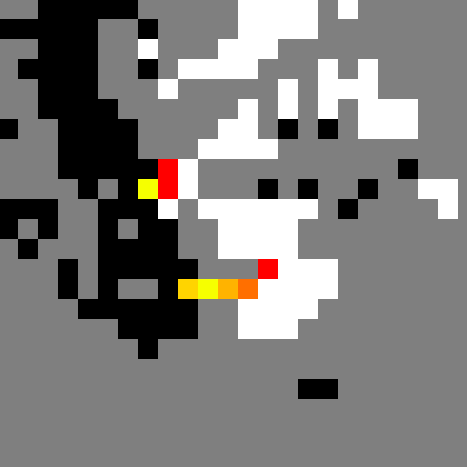}
\end{tabular}    
    \caption{{\bf Predicting keypoints. } 
    When events are integrated over a long time window, edges become thick and accurate keypoint localization becomes challenging. We therefore predict the keypoint's spatial position for multiple time steps. The color gradient from yellow to red represents the time from past to present. Our network is able to predict precise keypoint locations at every time step and predict their trajectory through time. (Best seen in colour)
    }
    \label{fig:corners_on_synthetic}
\end{figure*}

\section{Experiments}
\label{section:results}

In this section, we report our experimental comparison with previous event-based keypoint detectors, and an ablation study on the different aspects of our detector.
We first introduce the datasets and metrics we consider, and then report the results of our experiments.

\subsection{Datasets and Metrics}

We consider the HVGA ATIS Corner dataset~\cite{manderscheid2019speed} build to evaluate event-based keypoint detectors. It consists of seven sequences with varying degrees of texture from a standard checkerboard to a complex natural image. These sequences were captured using an ATIS sensor with a resolution of $480\times360$ pixels. It only contains recordings of planar patterns.

Following the evaluation protocol defined in \cite{manderscheid2019speed, gradients}, we consider the following metrics:
\begin{itemize}
    \item the $\delta t$-homography reprojection error $\text{er}_{\delta t}$, which depends on a time parameter $\delta t$: 
    \begin{equation}
        \text{er}_{\delta t}= \frac{1}{N_K} \sum_{t} \sum_k \lVert W_{t}^{t+\delta t}(K_{t, k}) - K_{t+\delta t, k} \lVert \> ,
    \end{equation}
     where $K_{t, k}$ is a detected keypoint at time $t$, and $K_{t+\delta t, k}$ is the location of the keypoint at time $t+\delta t$ that belongs to the same track as keypoint $K_{t, k}$. If the track of $K_{t, k}$ ends before $t+\delta t$, it is ignored in the sum. $W_{t}^{t+\delta t}$ is the estimated homography (using $K_{t, k}$ and $K_{t+\delta t, k}$) that warps the view of the planar scene at time $t$ to the one at time $t+\delta t$. $N_K$ is the total number of terms in the sums, to compute the average of the distances.
    \item the average track lifetime of the longest 100 tracks over each of the seven sequence.
\end{itemize}

\begin{figure*}[htbp]
\centering
\begin{tabular}{c@{ }c@{ }c}
    \includegraphics[width=.33\linewidth]{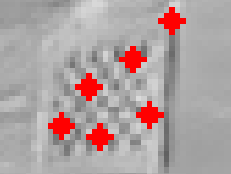}&
    \includegraphics[width=.33\linewidth]{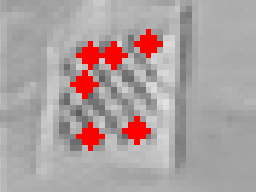}&
    \includegraphics[width=.33\linewidth]{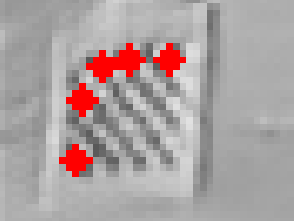}
\end{tabular}    
    \caption{{\bf Ground truth inconsistencies for the Event-Camera Dataset~\cite{zurich_event_dataset}. } Detecting ground-truth keypoint as Harris corners from consecutive event-to-video frames leads to unstable labels. (Best seen in colour)}
    \label{fig:ground_truth_errors}
\end{figure*}

We also evaluated our detector on The Event-Camera Dataset and Simulator~\cite{zurich_event_dataset}. The dataset is composed of multiple scenes with a variety of shapes and textures. Some scenes capture 3D and/or dynamic environments. This makes the computation of the homography reprojection error metric unfeasible. We therefore followed a similar protocol to \cite{luvharris} for evaluation: We run the event-to-video model from \cite{e2vid} to recreate frames from events and extract Harris corners on said frames to create a groundtruth. However, as can be seen in Figure~\ref{fig:ground_truth_errors}, the groundtruth is inaccurate and unstable with time. Another flaw of this evaluation criterion is a positive bias towards keypoints detectors using, or based on, Harris corners. Nonetheless,  for completeness, we provide an evaluation following this protocol for comparison with earlier methods as it was used in previous papers.

On this dataset, to be able to compare with already published results, we report $P_\text{rel}$ and 
$R_\text{rel}$, respectively the precision and recall relative to eHarris. As in \cite{luvharris} results are computed on the sequences \texttt{boxes\_6dof}, \texttt{dynamic\_6dof}, \texttt{poster\_6dof} and \texttt{shapes\_6dof}.

\subsection{Ablation Study}

\begin{table}[htbp]
\begin{center}
\begin{tabular}{c@{$\quad\quad\quad$}c@{$\quad$}c@{$\quad$}c@{$\quad$}c@{$\quad$}c@{$\quad\quad$}c}
 \toprule
  \multirow{2}{*}{$N_H$} & \multicolumn{5}{c}{$\delta t$-reprojection Error (pixels)$\quad\quad$} & \multirow{2}{*}{Track Lifetime (s)}\\
  & $\delta t=$ 25 & 50 & 100 & 150 & 200 & \\
 \hline
 1 & 1.47 & 1.62 & 1.91 & 1.95 & 2.62 & 15.63 \\
 2 & 1.27 & 1.47 & 1.66 & 1.81 & 2.06 & \textbf{16.70} \\
 3 & 1.25 & 1.32  & 1.60 & 1.67 & 1.79 & 16.44 \\
 5 & 1.26 & 1.31 & 1.47 & 1.82 & 1.85 & 16.14 \\
 10 & 1.18 & 1.28 & 1.45 & 1.63 & 1.84 & 15.7 \\
 12 & \textbf{1.15} & \textbf{1.24} & \textbf{1.35} & \textbf{1.53} & \textbf{1.68} & 15.4 \\

 \bottomrule
\end{tabular}
\end{center}
\caption{{\bf Influence of $N_H$, the number of predicted heatmaps.} $N_H = 1$ corresponds to the standard approach with a single prediction for the entire integration period. While $N_H = 1$ already performs well
thanks to our training procedure, increasing $N_H$ consistently improves accuracy.   \label{table:multi_prediction}}
\end{table}

\paragraph{Predicting a single heatmap vs multiple heatmaps, and the influence of $N_H$.}
We evaluated the performance of our detector on the HVGA ATIS Corner Dataset when varying $N_H$, the number of predicted heatmaps. Setting $N_H = 1$ allows us to also evaluate the influence of predicting multiple heatmaps compared to a single one, as it is more traditionally done.

Table~\ref{table:multi_prediction} reports the results. Predicting a single heatmap already performs well thanks to our training data. However, accuracy keeps improving when $N_H$ increases. 
As computation time also increases with $N_H$, there is still a balance to find. We chose $N_{H}=10$ for all the other experiments reported in this paper.

\paragraph{Influence of $\Delta T$, the length of integration period.} Our network is dependent on the integration period. Table~\ref{table:multi_delta_t} reports the $\delta t$-reprojection error and the track liftetime metric for various values of $\Delta T$.  Reducing $\Delta T$ too much reduces the performance as not enough events are given as inputs, which puts too much strain on the memory of such a small network. On the other hand, setting $\Delta T$ too high also reduces the precision of the predictions. A good trade-off is $\Delta t = 5ms$, and it is the value we used for all other experiments. 

\begin{table}[htbp]
\begin{center}
\begin{tabular}{c@{$\quad\quad\quad$}c@{$\quad$}c@{$\quad$}c@{$\quad$}c@{$\quad$}c@{$\quad\quad$}c}
 \hline
  \multirow{2}{*}{Integration Period $\Delta T$} & \multicolumn{5}{c}{$\delta t$-reprojection Error (pixels)$\quad\quad$} & \multirow{2}{*}{Track Lifetime (s)}\\
  & $\delta t=$ 25 & 50 & 100 & 150 & 200 & \\
 \hline
 2.5ms & 1.19 & 1.33 & 1.72 & 2.15 & 3.11 & 12.8 \\
 5ms & \textbf{1.18} & \textbf{1.28} & \textbf{1.45} & 1.63 & 1.84 & 15.7 \\
 7.5ms & 1.31 & 1.37 & 1.52 & \textbf{1.52} & \textbf{1.54} & \textbf{15.8} \\
 10ms & 1.39 & 1.50 & 1.51 & 1.55 & 1.56 & 14.8 \\
 25ms & 1.46 & 1.50 & 1.62 & 1.98 & 2.86 & 11.7 \\
 50 ms & - & 1.92 & 2.28 & 3.32 & 3.81 & 10.8 \\
 \hline
\end{tabular}
\end{center}
\caption{{\bf Influence of $\Delta T$, the length of integration period. }
    We use $\Delta T =$ 5ms in all the other experiments. \label{table:multi_delta_t}}
\end{table}



\begin{table}[htbp]
\begin{center}
\begin{tabular}{c@{$\quad\quad$}c@{$\quad$}c@{$\quad$}c@{$\quad$}c@{$\quad$}c@{$\quad$}c}
 \hline
  \multirow{2}{*}{Generation of Training Data} & \multicolumn{5}{c}{$\delta t$-reprojection Error (pixels)$\quad\quad$} & \multirow{2}{*}{Track Lifetime (s)}\\
  & $\delta t=$ 25 & 50 & 100 & 150 & 200 & \\
 \hline
 Detecting Corners & 1.34 & 1.59 & 1.99 & 2.19 & 2.31 & 2.4 \\
 Warping Corners & \textbf{1.18} & \textbf{1.28} & \textbf{1.45} & \textbf{1.63} & \textbf{1.84} & \textbf{15.7} \\
 \hline
\end{tabular}
\end{center}
\caption{{\bf Influence of detecting or warping keypoints during training. }
    We compare two networks: {\it Detecting Corners} corresponds to a network trained with keypoints detected independently in each frame;   {\it Warping Corners} corresponds to training with keypoints detected in the reference frame and then warped using the simulated homographies, as explained in Sec.~\ref{subsec:training_data}. \label{table:detect_vs_warp}}
\end{table}

\subsection{Comparison to the State-of-the-Art}

We compare ourselves to \cite{eharris, efast, arc, luvharris, manderscheid2019speed, gradients}. We have only limited results for \cite{manderscheid2019speed, gradients} as the authors did not make their models available.

\begin{table}[htbp]
\begin{center}
\begin{tabular}{c@{$\quad\quad\quad$}c@{$\quad$}c@{$\quad$}c@{$\quad$}c@{$\quad$}c@{$\quad\quad$}c}
 \hline
  \multirow{2}{*}{Method} & \multicolumn{5}{c}{$\delta t$-reprojection Error (pixels)$\quad\quad$} & \multirow{2}{*}{Track Lifetime (s)}\\
  & $\delta t = 25$ & 50 & 100 & 150 & 200 & \\
 \hline
 eHarris~\cite{eharris} & 2.57 & 3.46 & 4.58 & 5.37 & 6.06 & 0.74 \\
 eFast~\cite{efast} & 2.12 & 2.63 & 3.18 & 3.57 & 3.82 & 0.69 \\
 Arc~\cite{arc}& 3.80 & 5.31 & 7.22 & 8.48 & 9.49 & 0.91 \\
 SILC~\cite{manderscheid2019speed}& 2.45 & 3.02 & 3.68 & 4.13 & 4.42 & 1.12 \\
 gradients~\cite{gradients}& 2.46 & - & - & - & - & 5.46 \\
 ours & \textbf{1.18} & \textbf{1.28} & \textbf{1.45} & \textbf{1.63} & \textbf{1.84} & \textbf{15.7} \\
 \hline
\end{tabular}
\end{center}
\caption{{\bf  Evaluation on the HVGA ATIS Corner Dataset.} Our detector outperforms all previous methods in terms of $\delta t$-homography reprojection error for various $\delta t$, and  of track lifetime metric. The reprojection errors for Gradients are not available for $\delta t > 25$. We nearly reduce by half the reprojection error while at the same time triple the tracks lifetime. \label{table:comparison_others}}
\end{table}

\begin{figure*}[t]
\centering
\begin{tabular}{ccccc}
    \includegraphics[width=.19\linewidth]{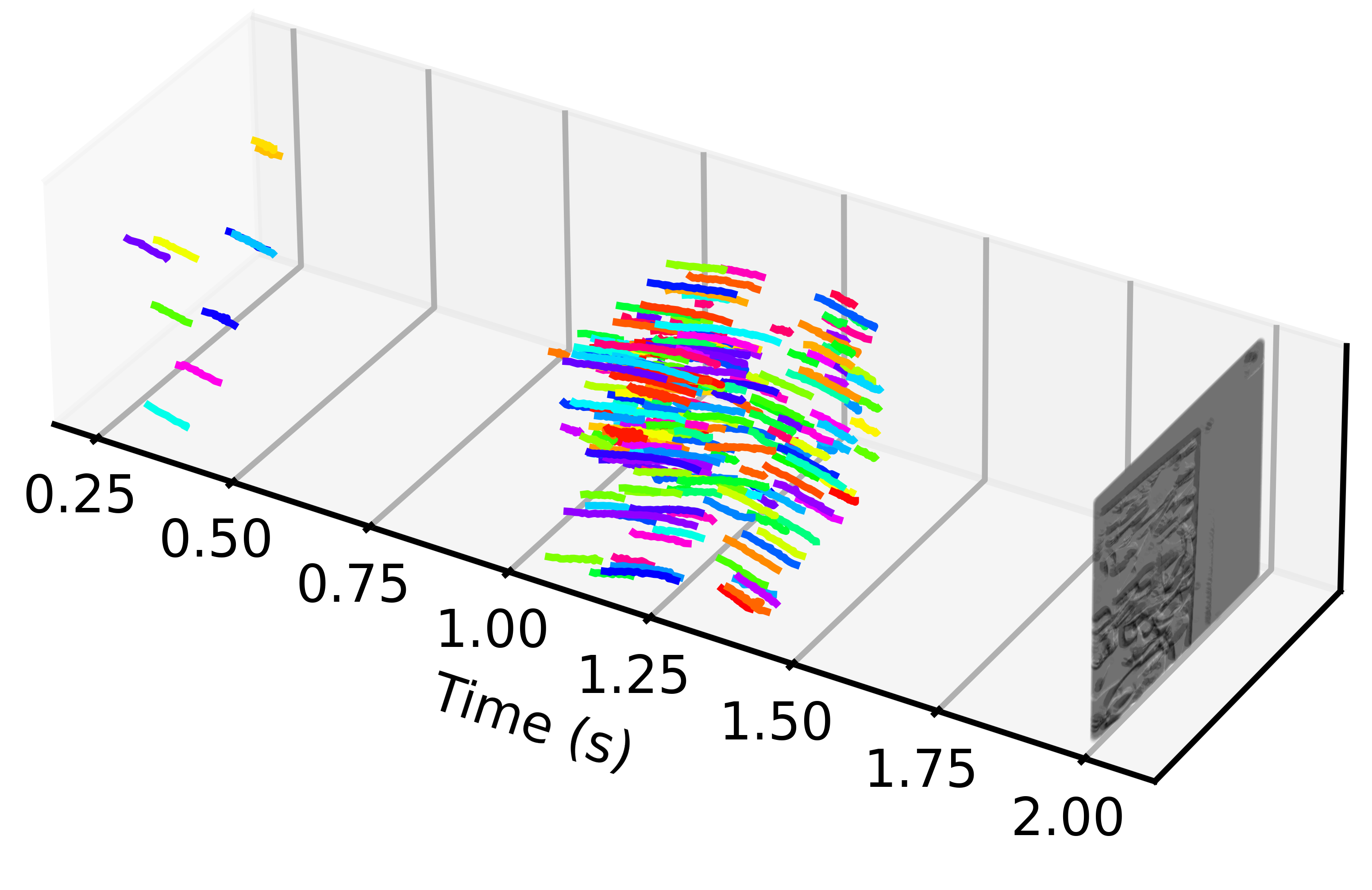}&
    \includegraphics[width=.19\linewidth]{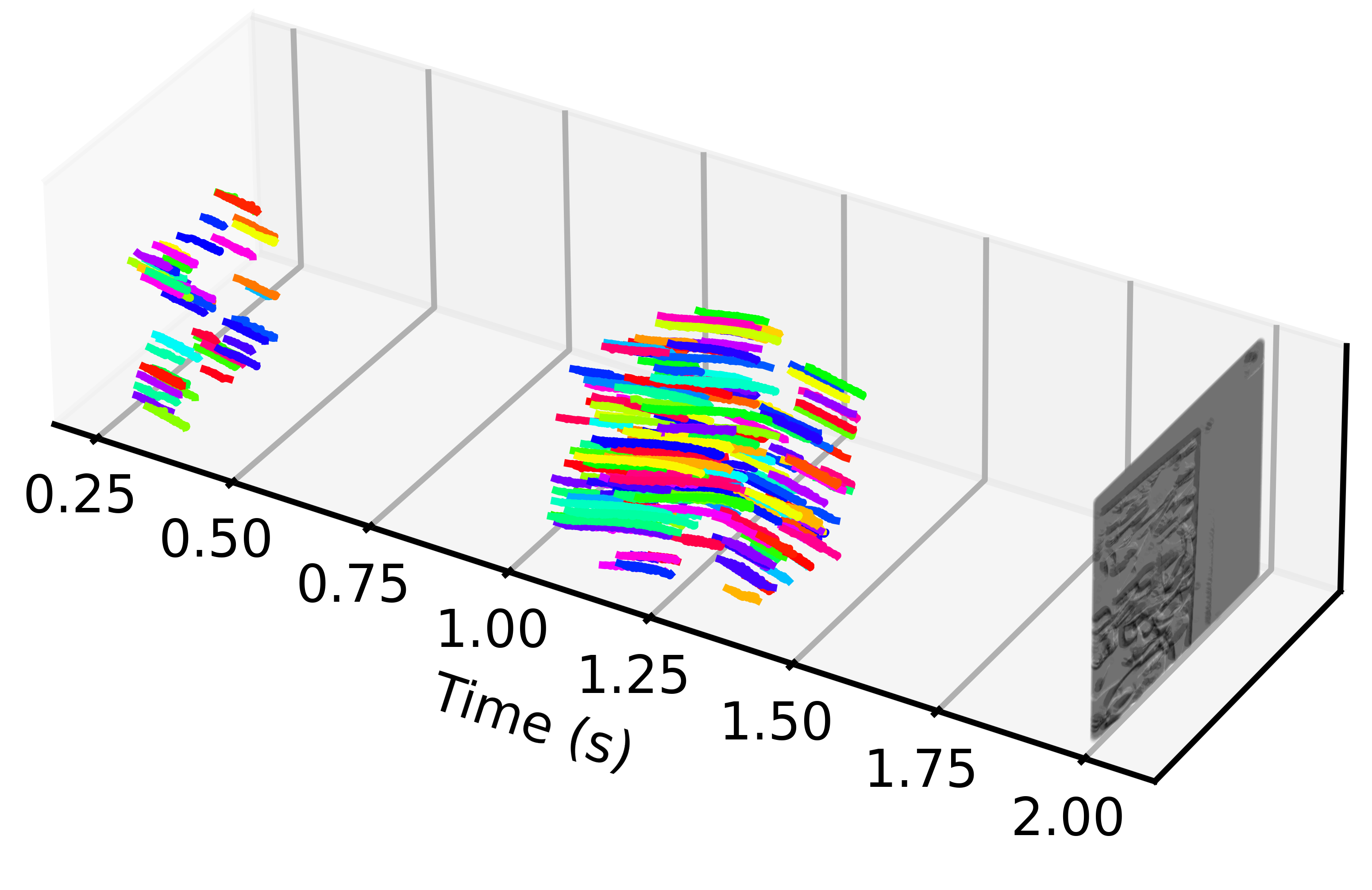}&
    \includegraphics[width=.19\linewidth]{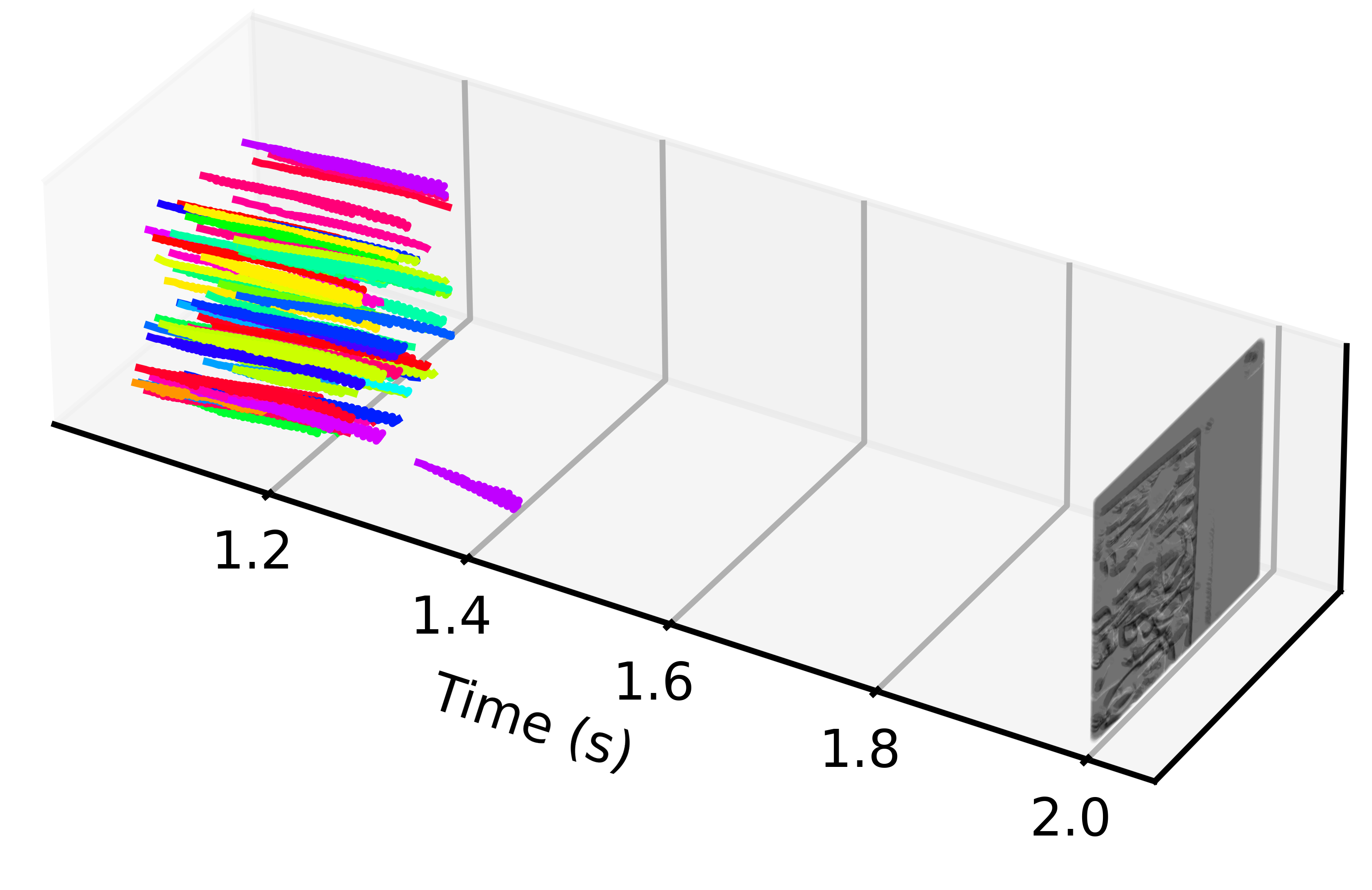}&
    \includegraphics[width=.19\linewidth]{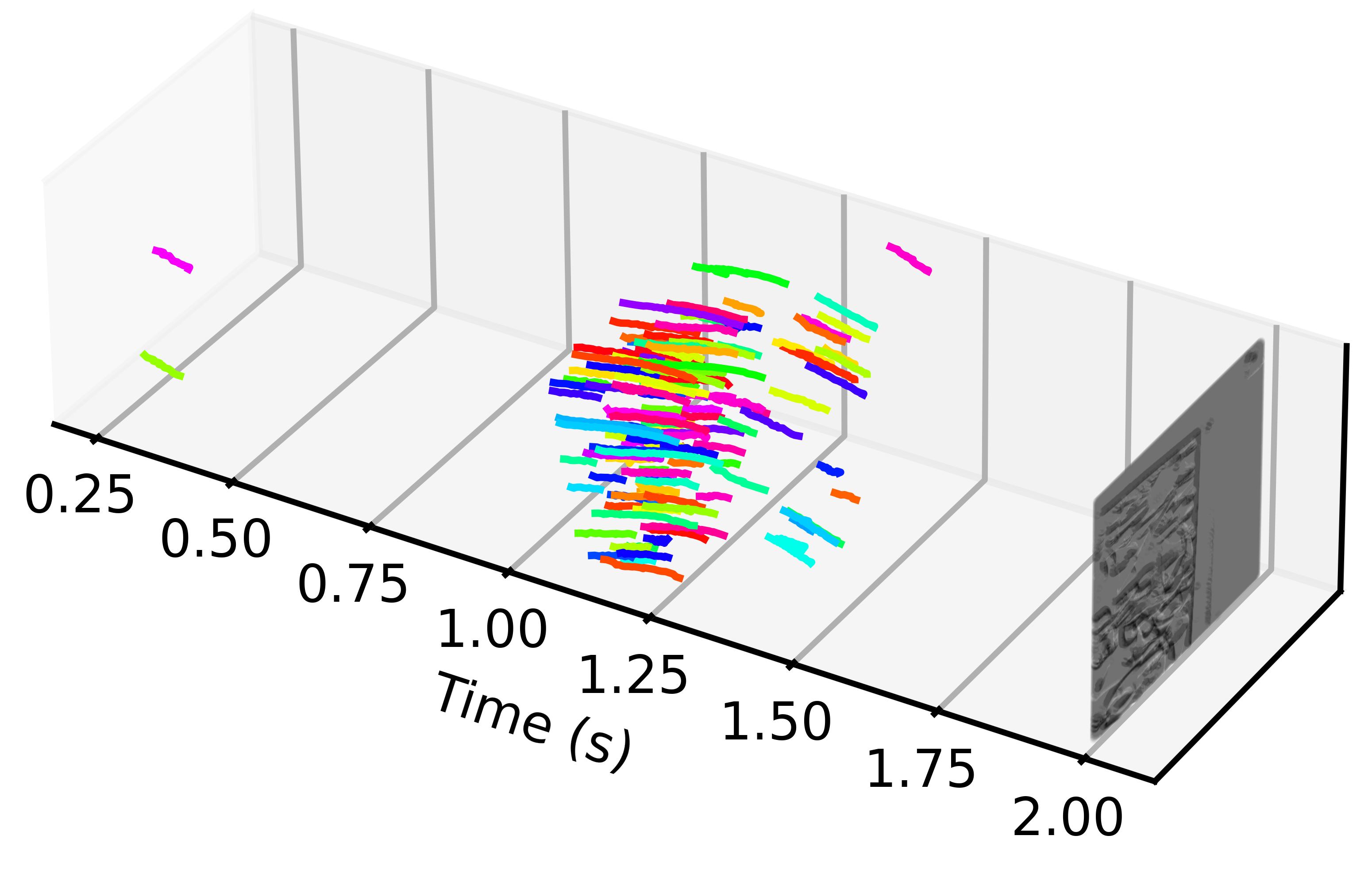}&
    \includegraphics[width=.19\linewidth]{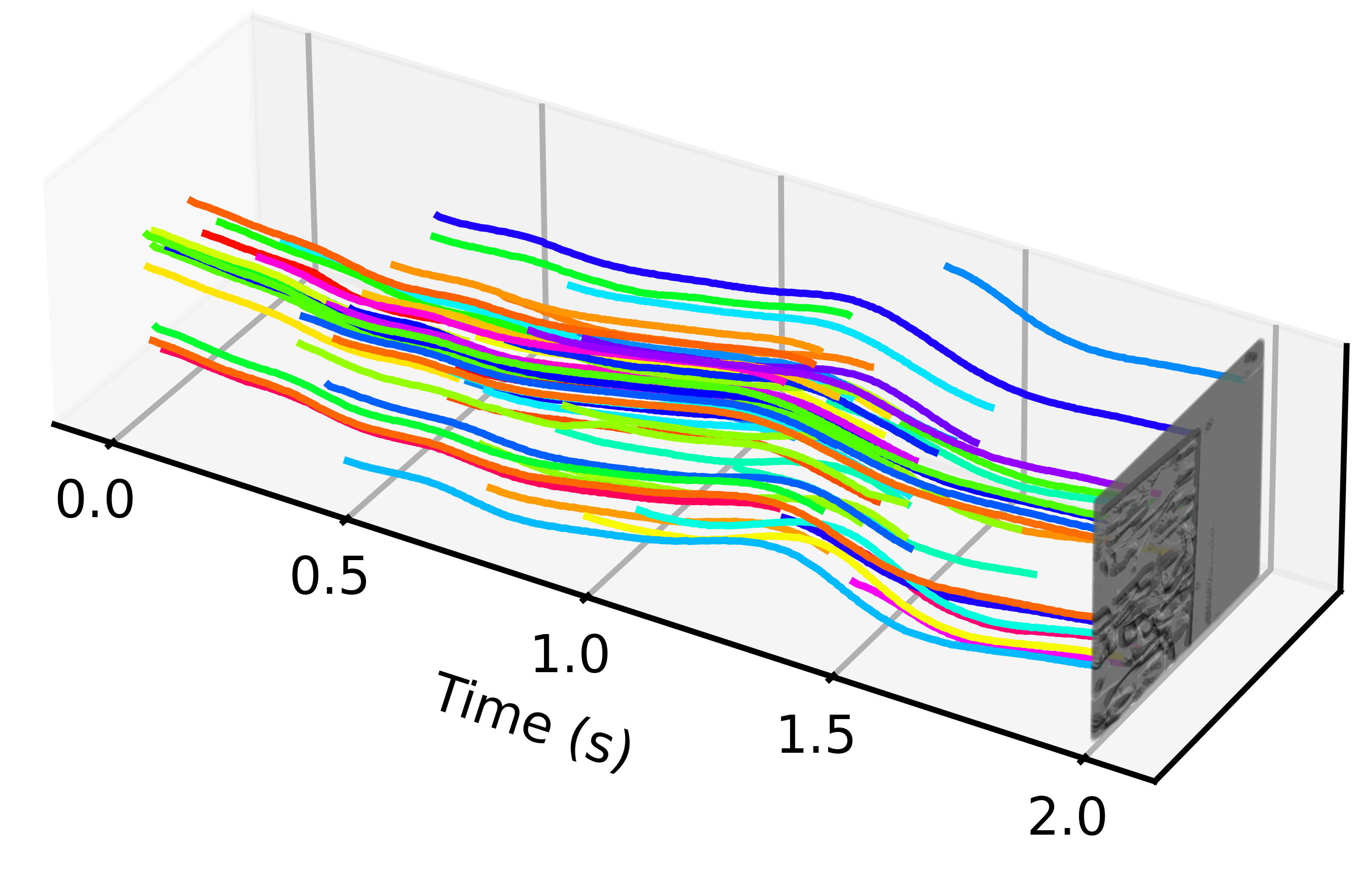}
    \\
    \includegraphics[width=.19\linewidth]{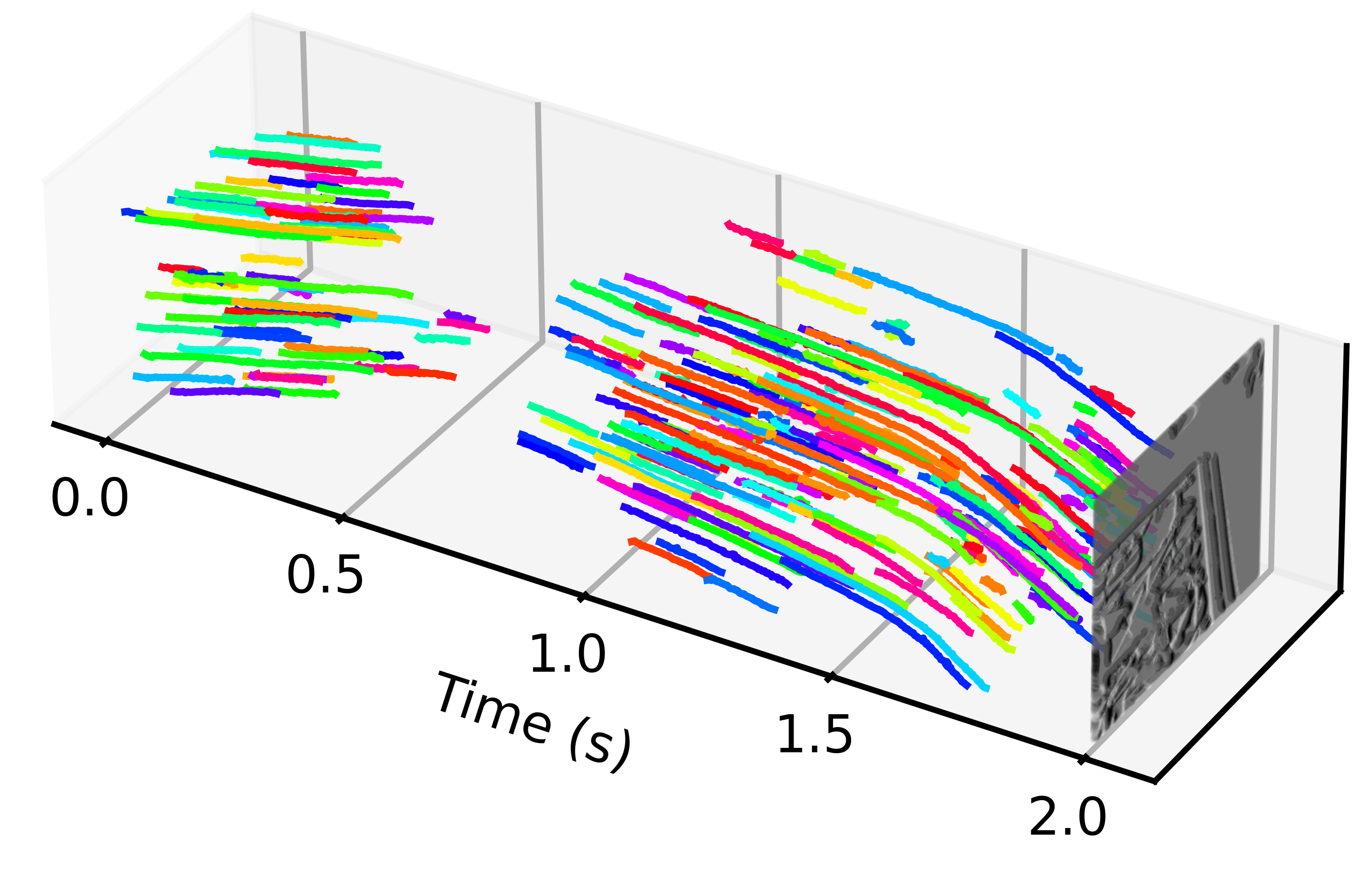}&
    \includegraphics[width=.19\linewidth]{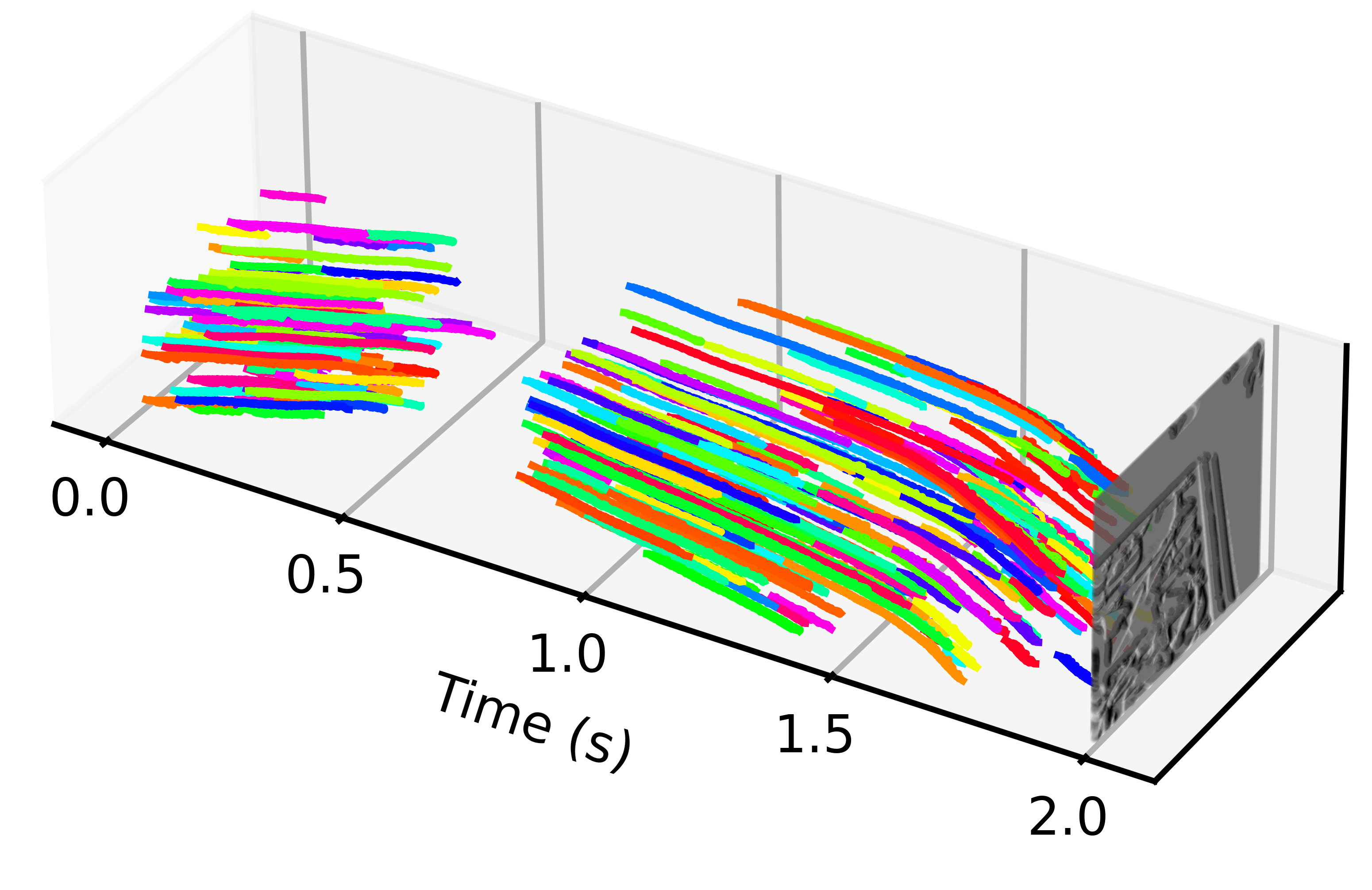}&
    \includegraphics[width=.19\linewidth]{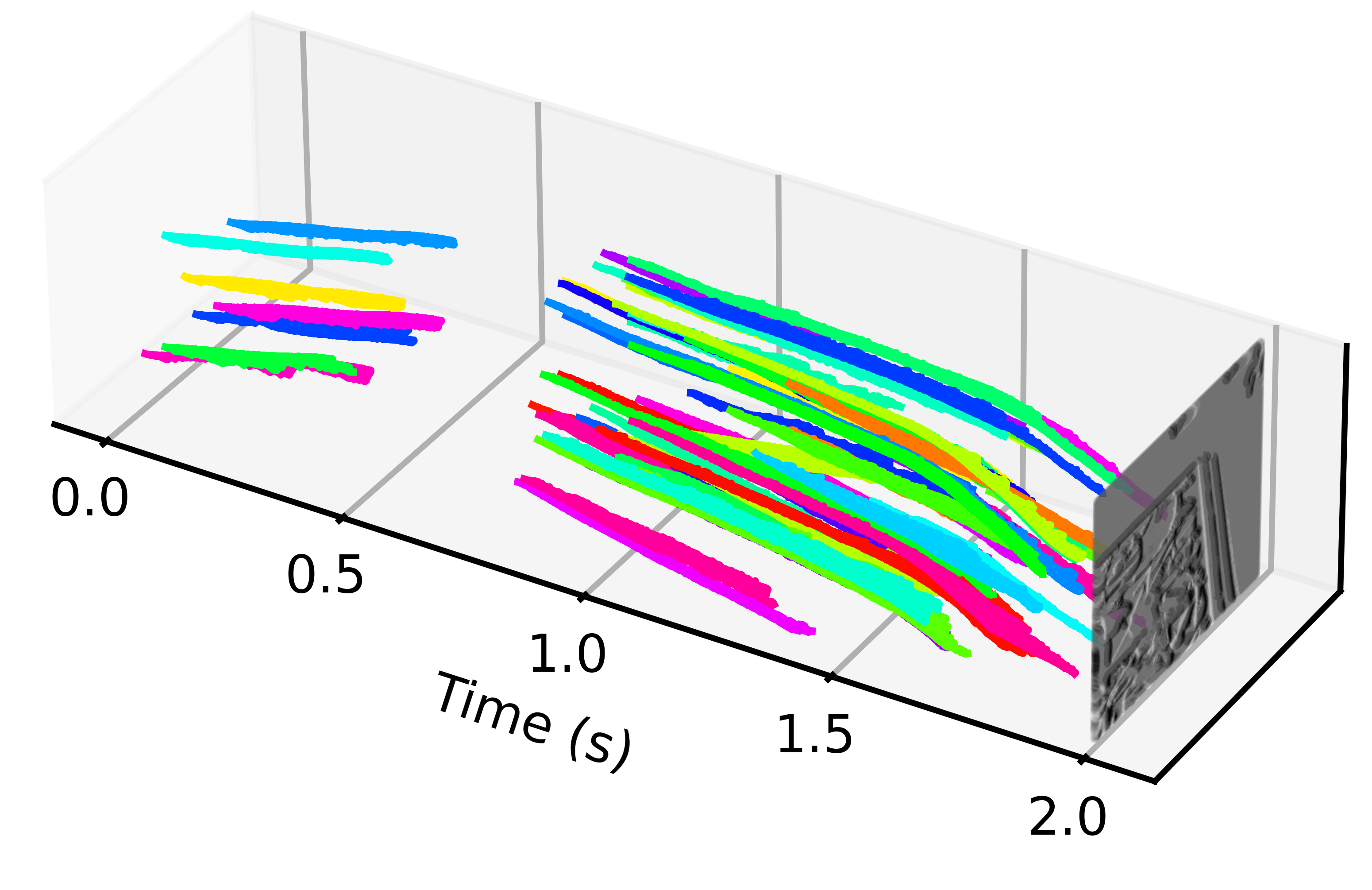}&
    \includegraphics[width=.19\linewidth]{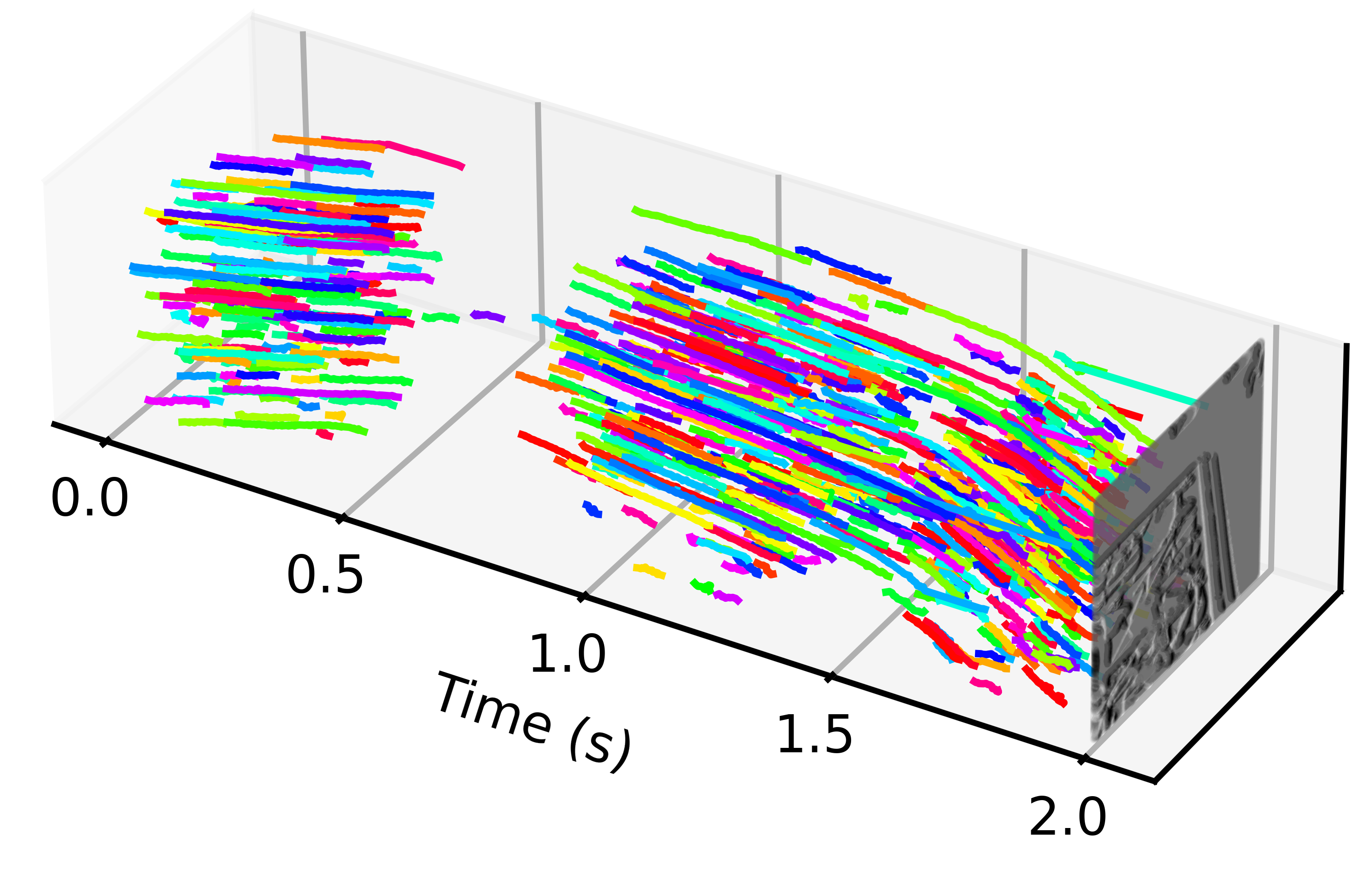}&
    \includegraphics[width=.19\linewidth]{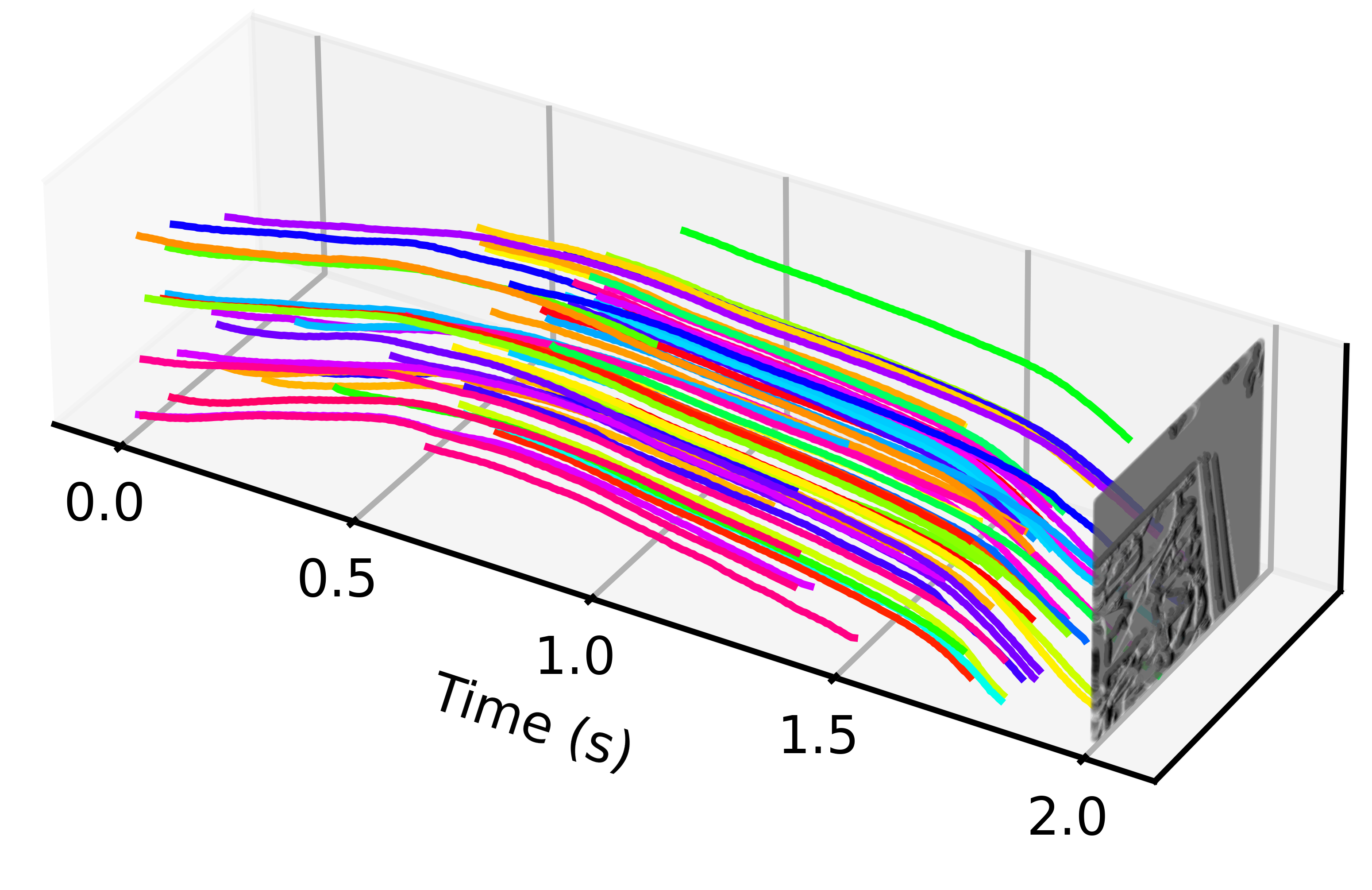}
    \\
    \includegraphics[width=.19\linewidth]{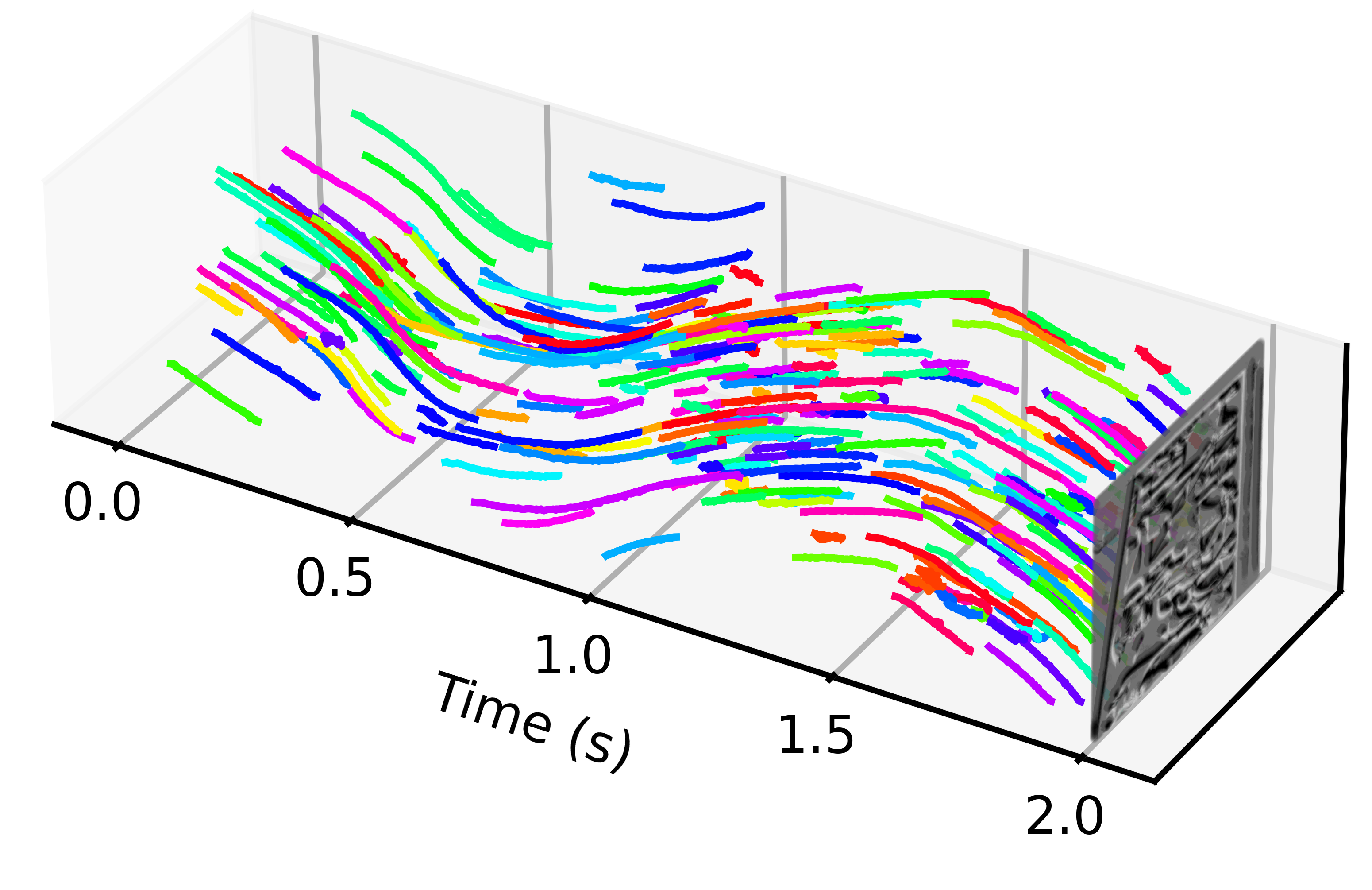}&
    \includegraphics[width=.19\linewidth]{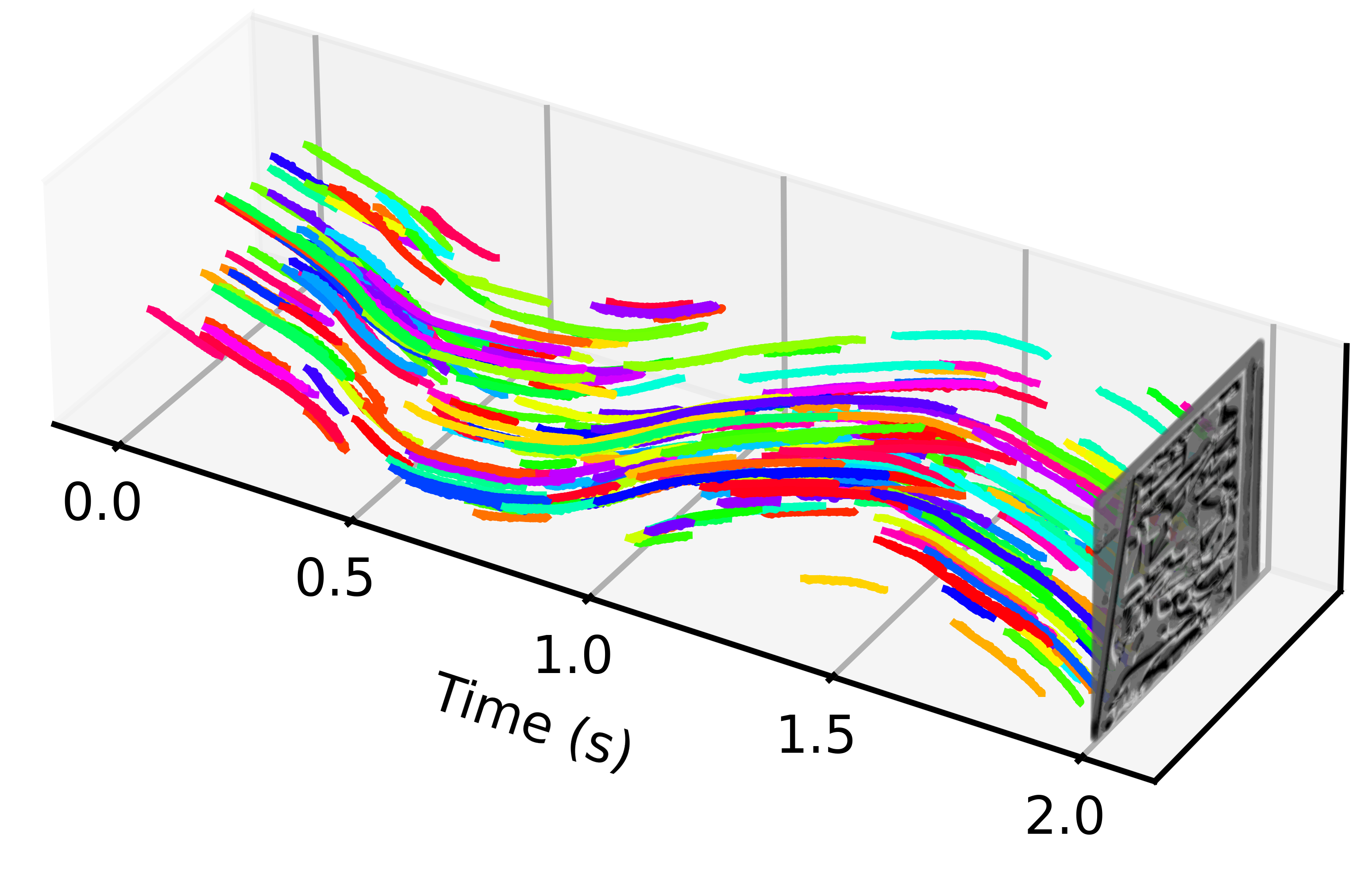}&
    \includegraphics[width=.19\linewidth]{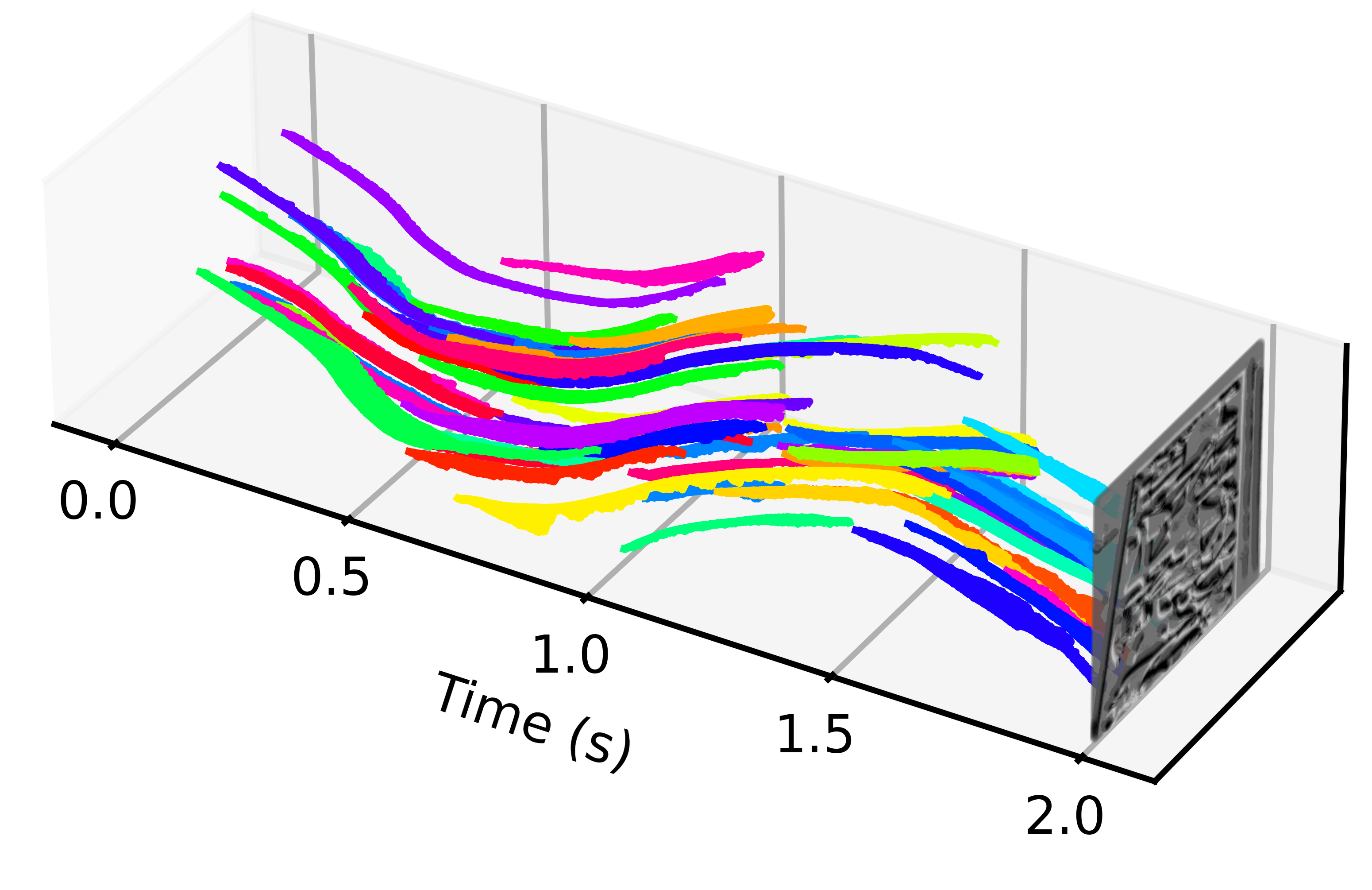}&
    \includegraphics[width=.19\linewidth]{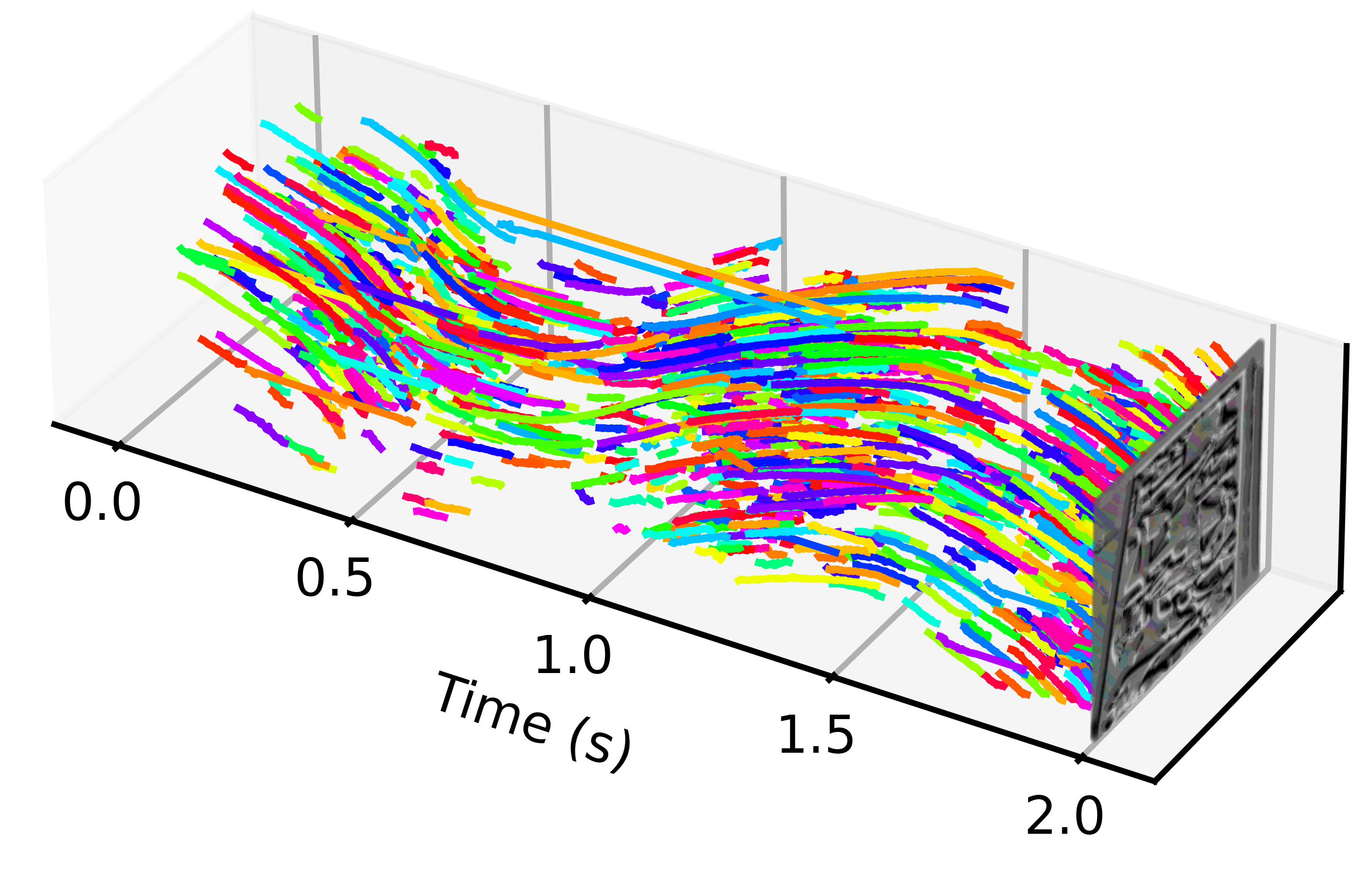}&
    \includegraphics[width=.19\linewidth]{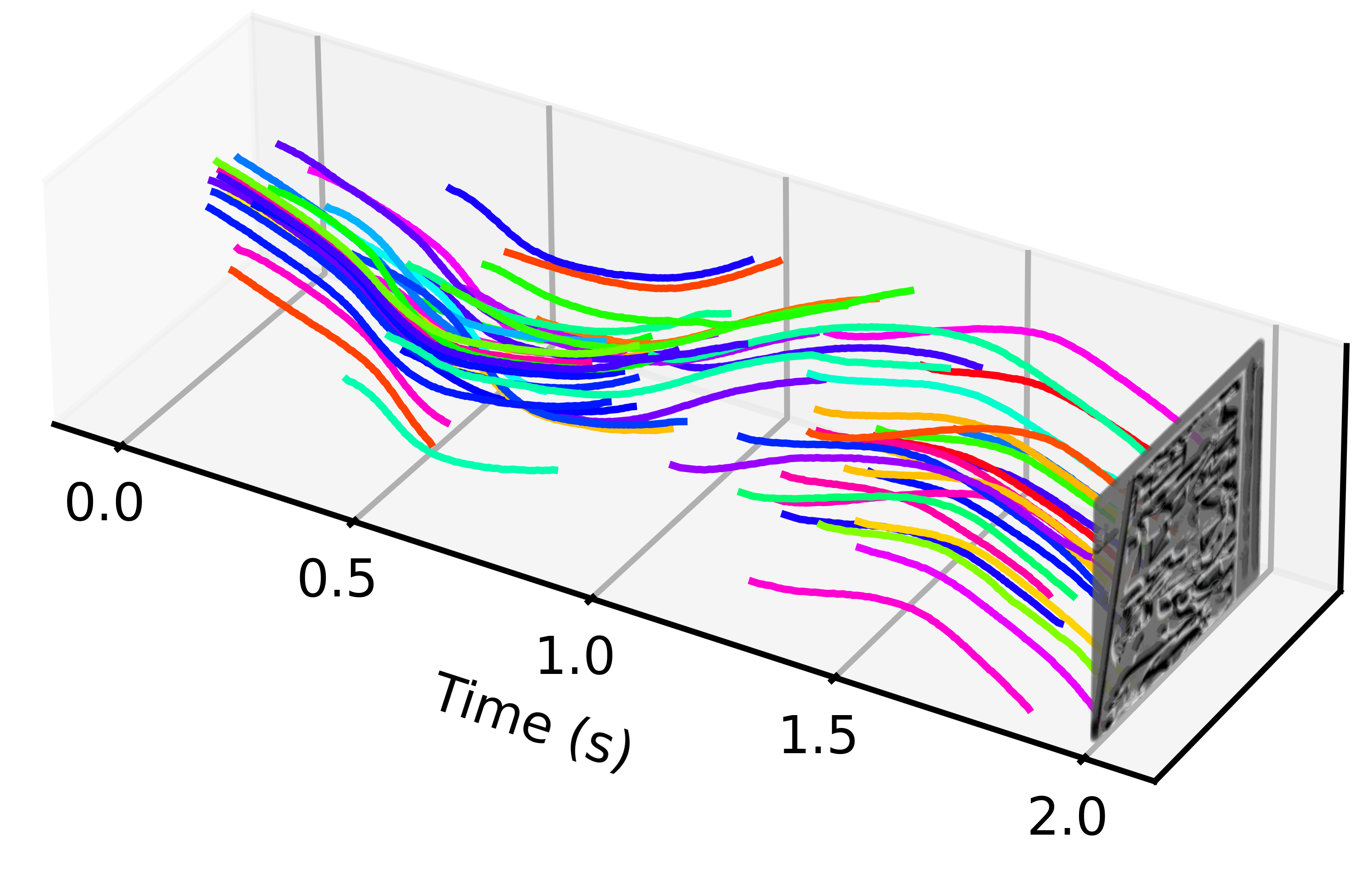}
    \\
    \includegraphics[width=.19\linewidth]{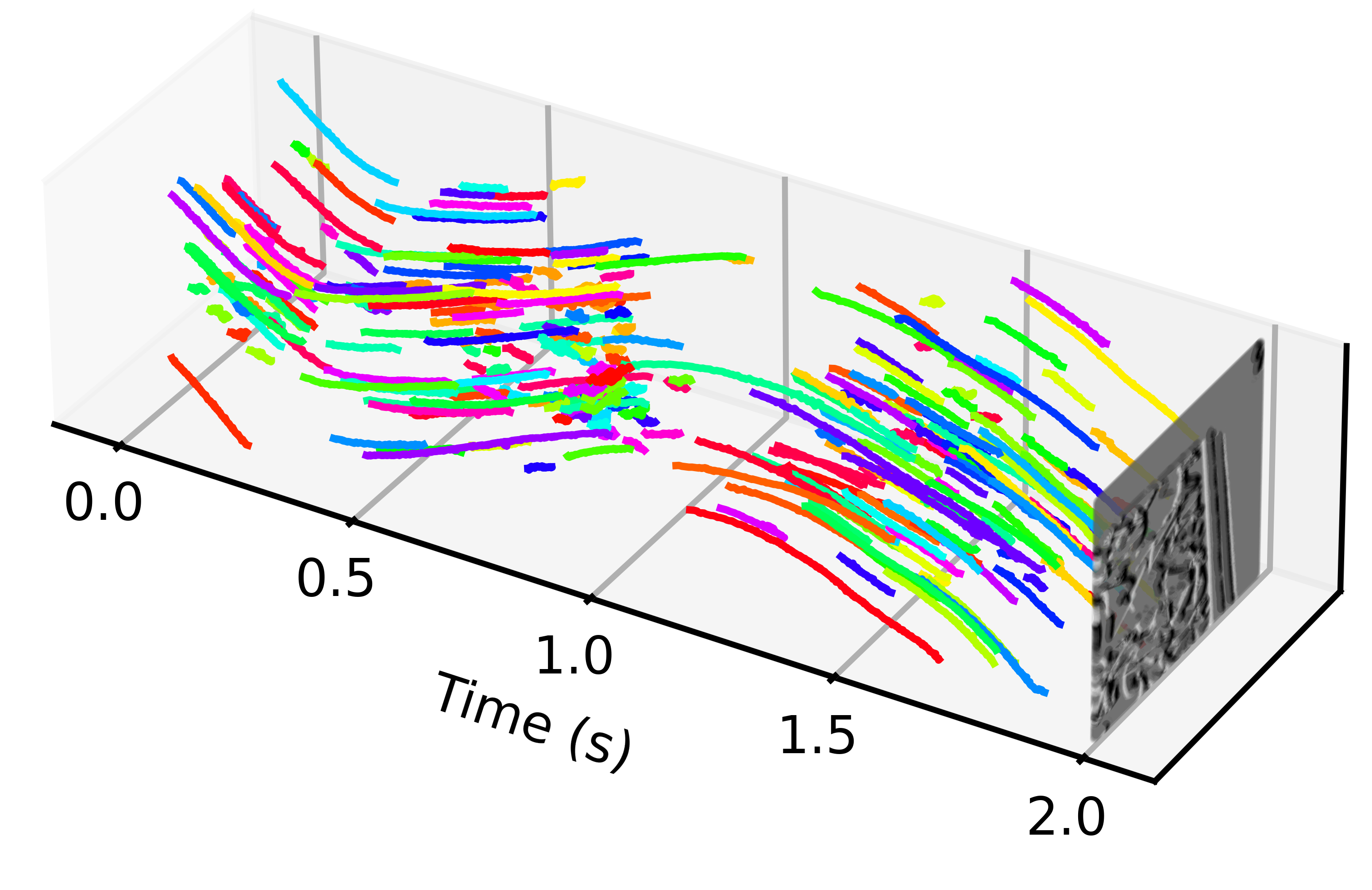}&
    \includegraphics[width=.19\linewidth]{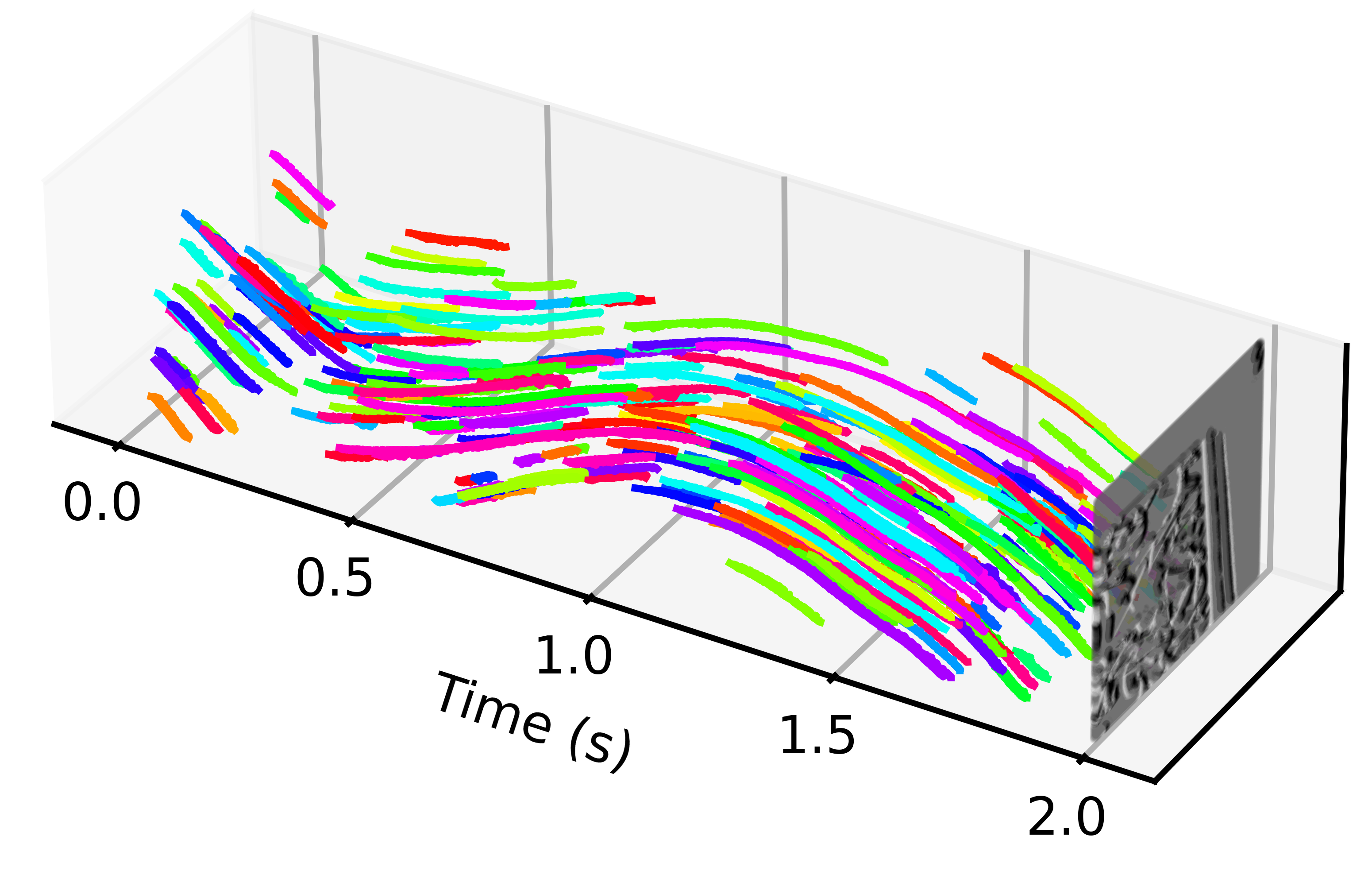}&
    \includegraphics[width=.19\linewidth]{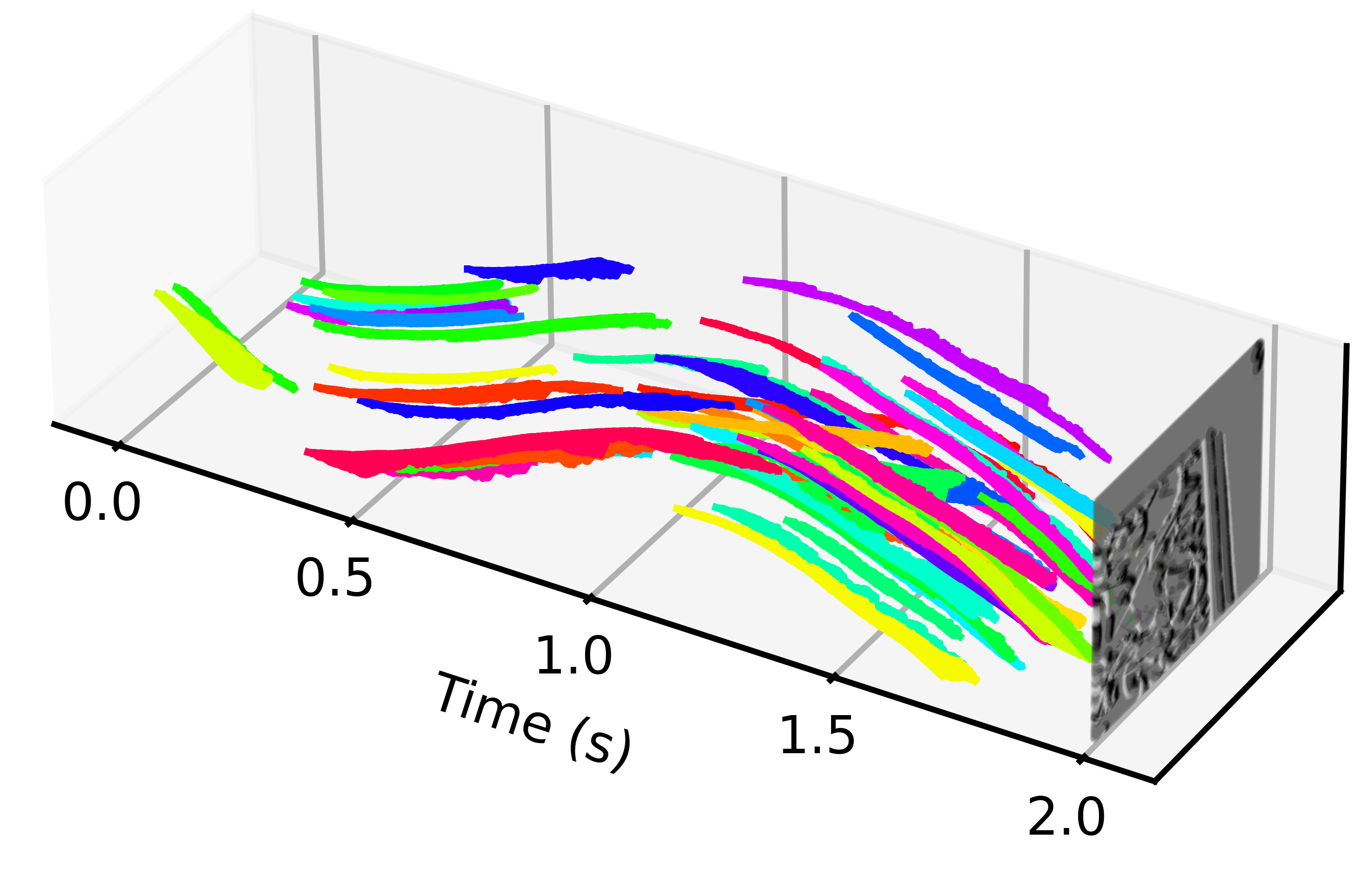}&
    \includegraphics[width=.19\linewidth]{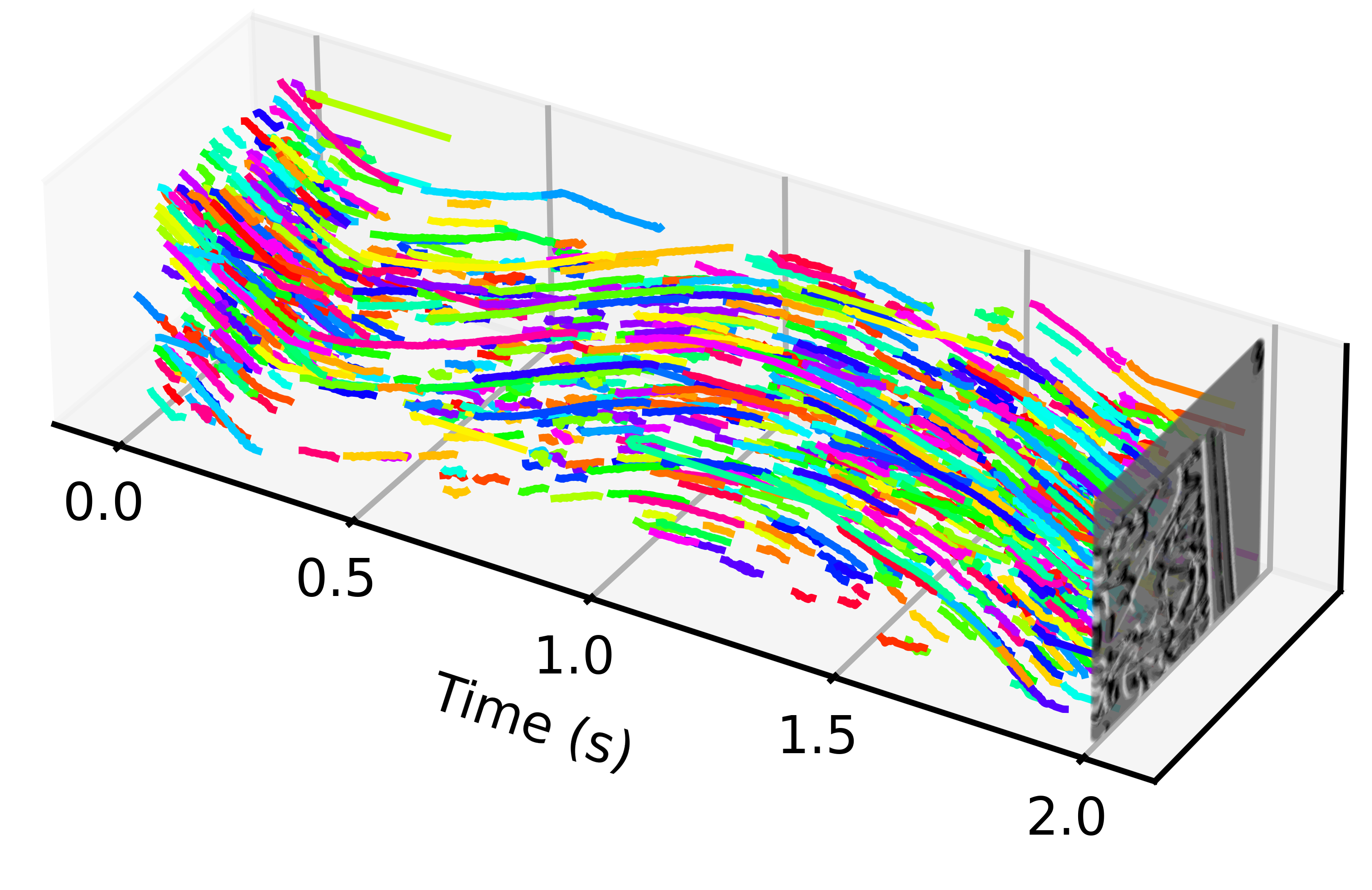}&
    \includegraphics[width=.19\linewidth]{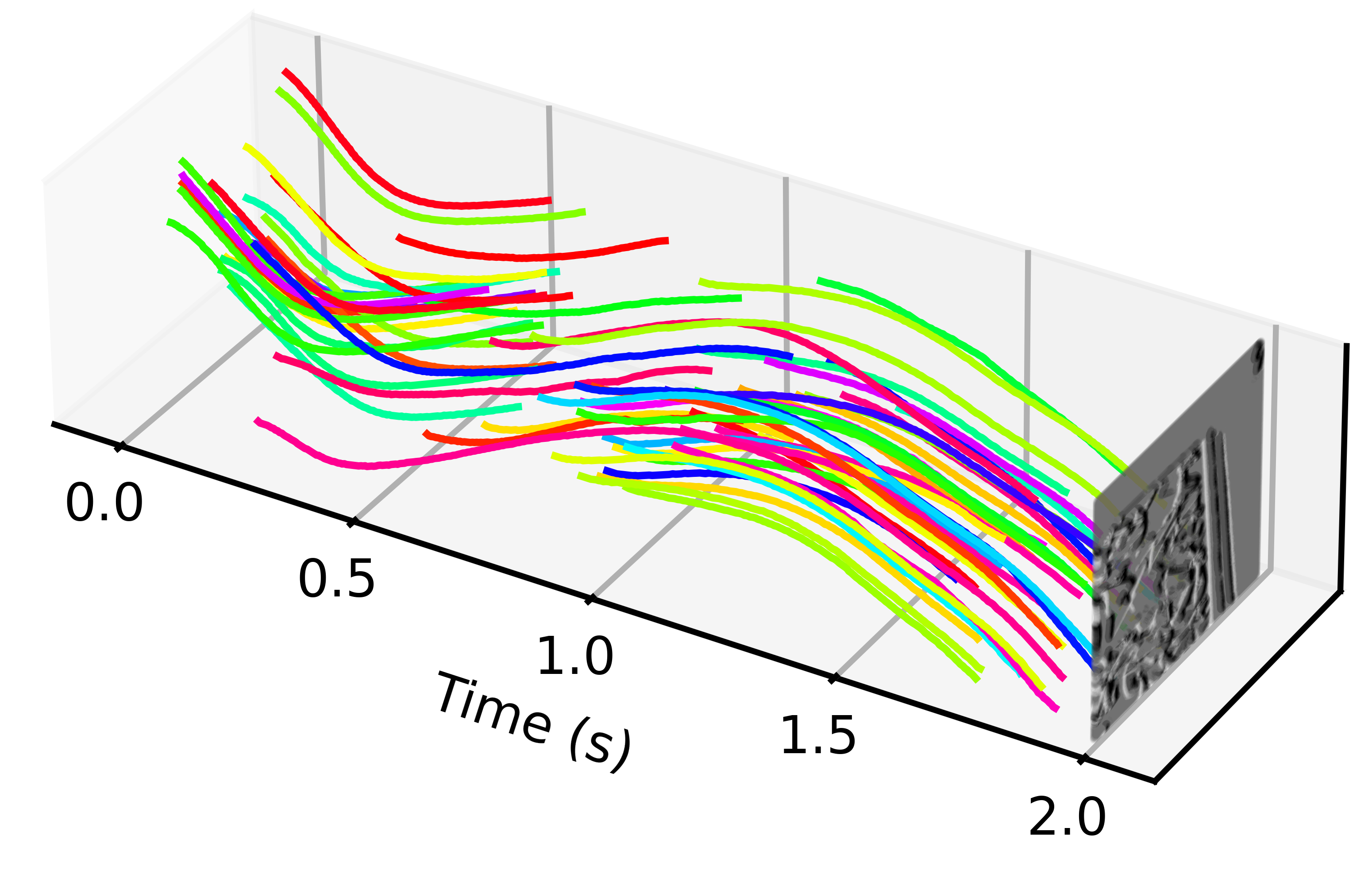}
    \\
    \includegraphics[width=.19\linewidth]{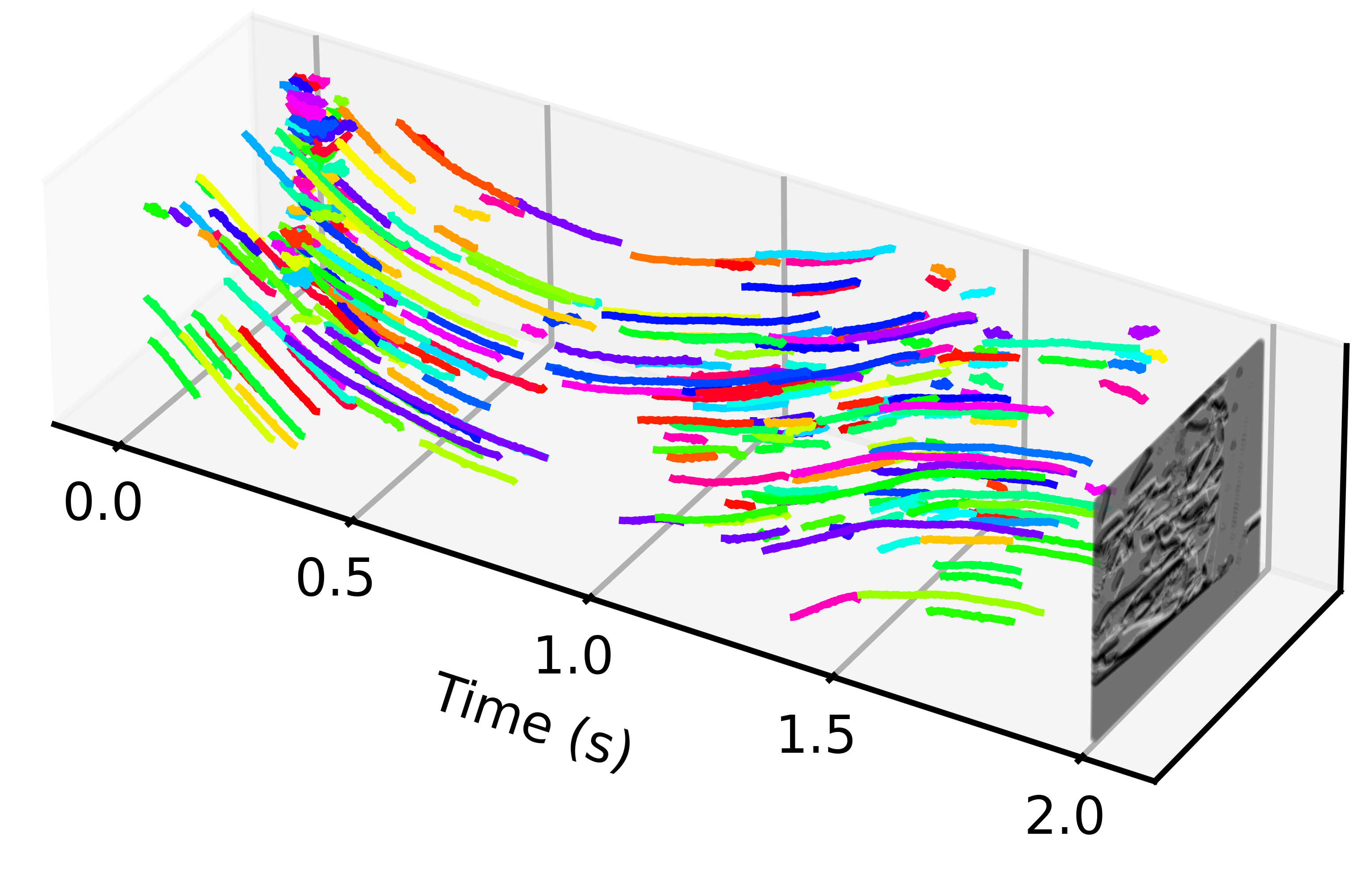}&
    \includegraphics[width=.19\linewidth]{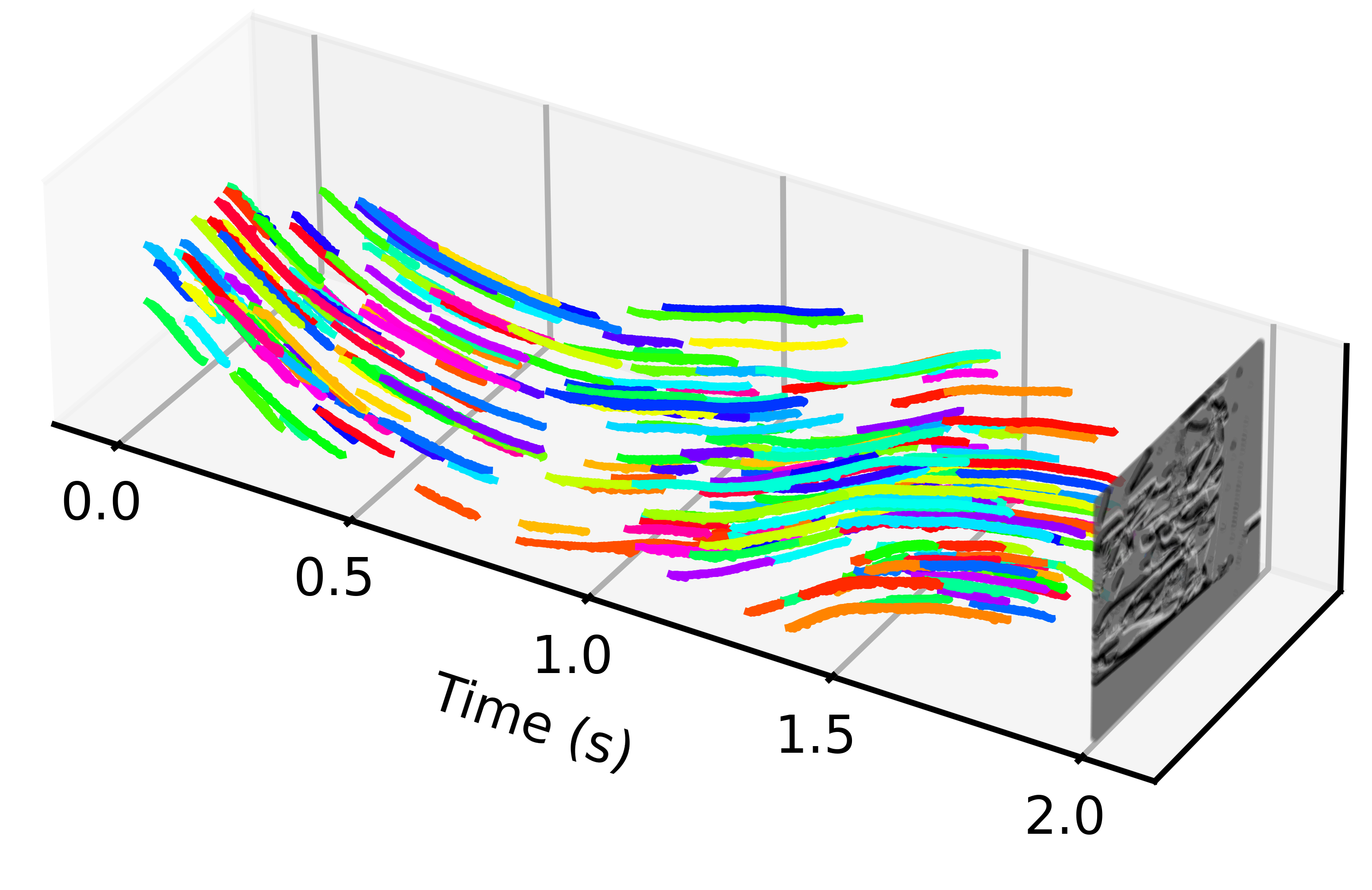}&
    \includegraphics[width=.19\linewidth]{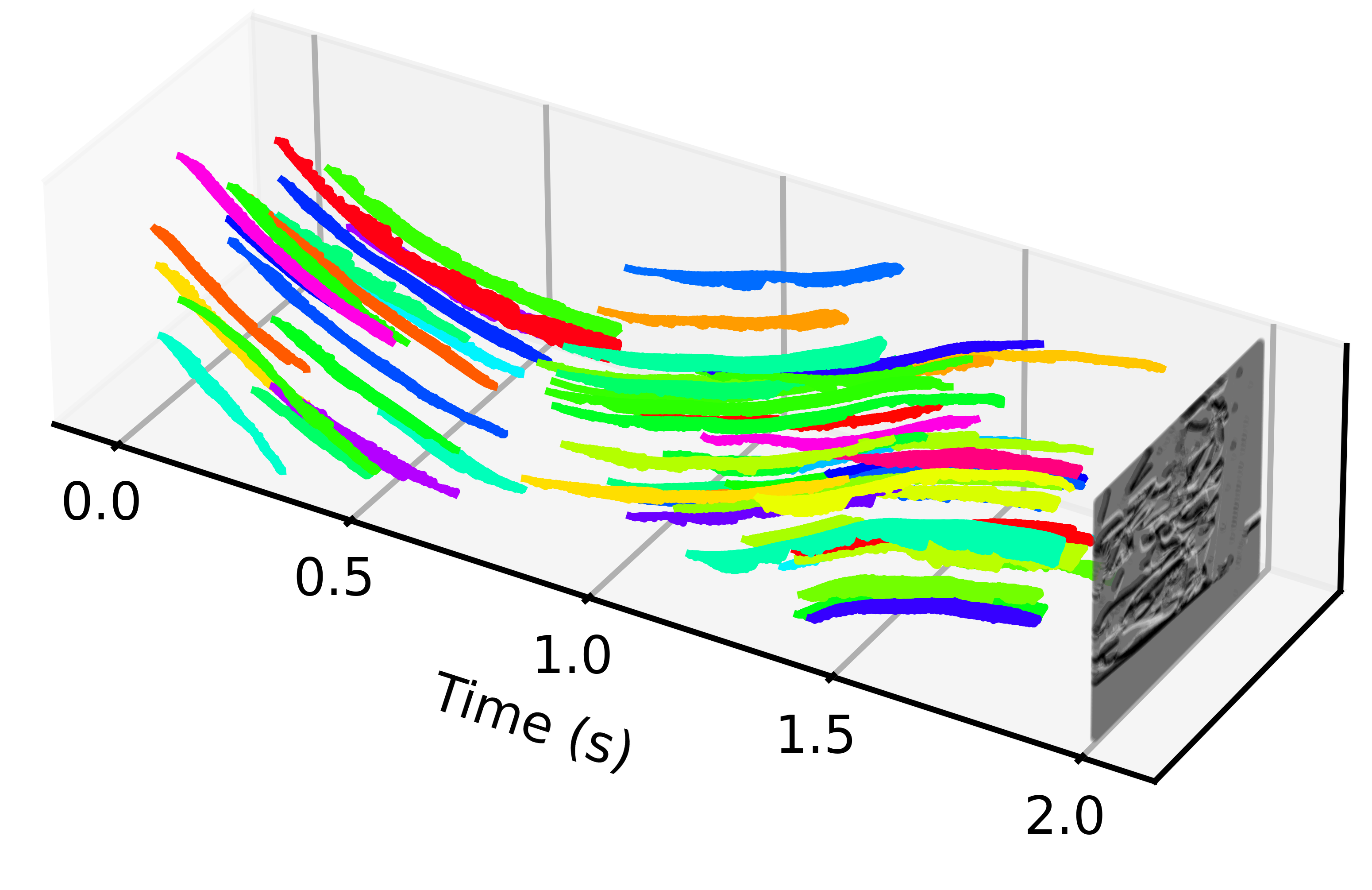}&
    \includegraphics[width=.19\linewidth]{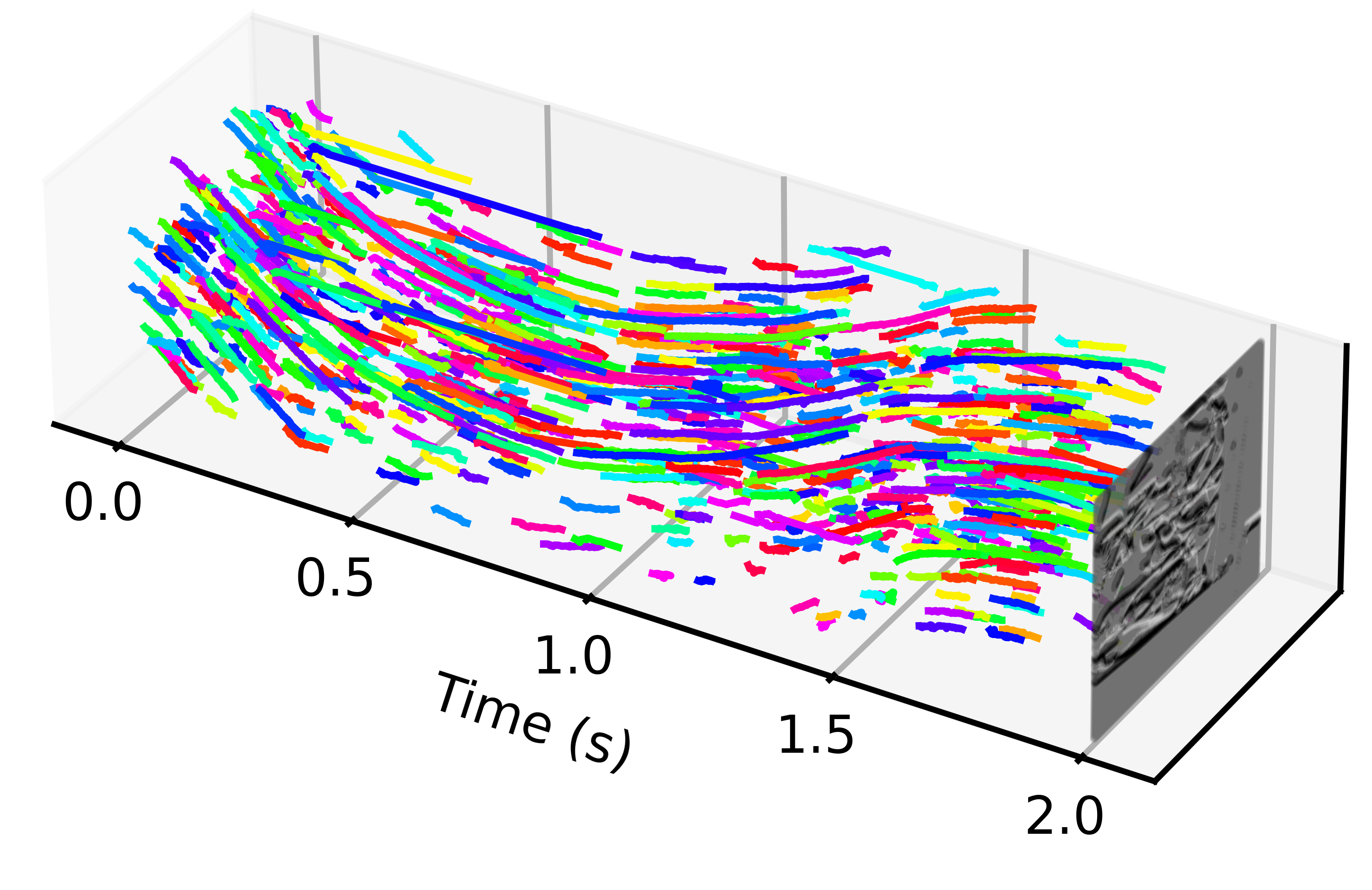}&
    \includegraphics[width=.19\linewidth]{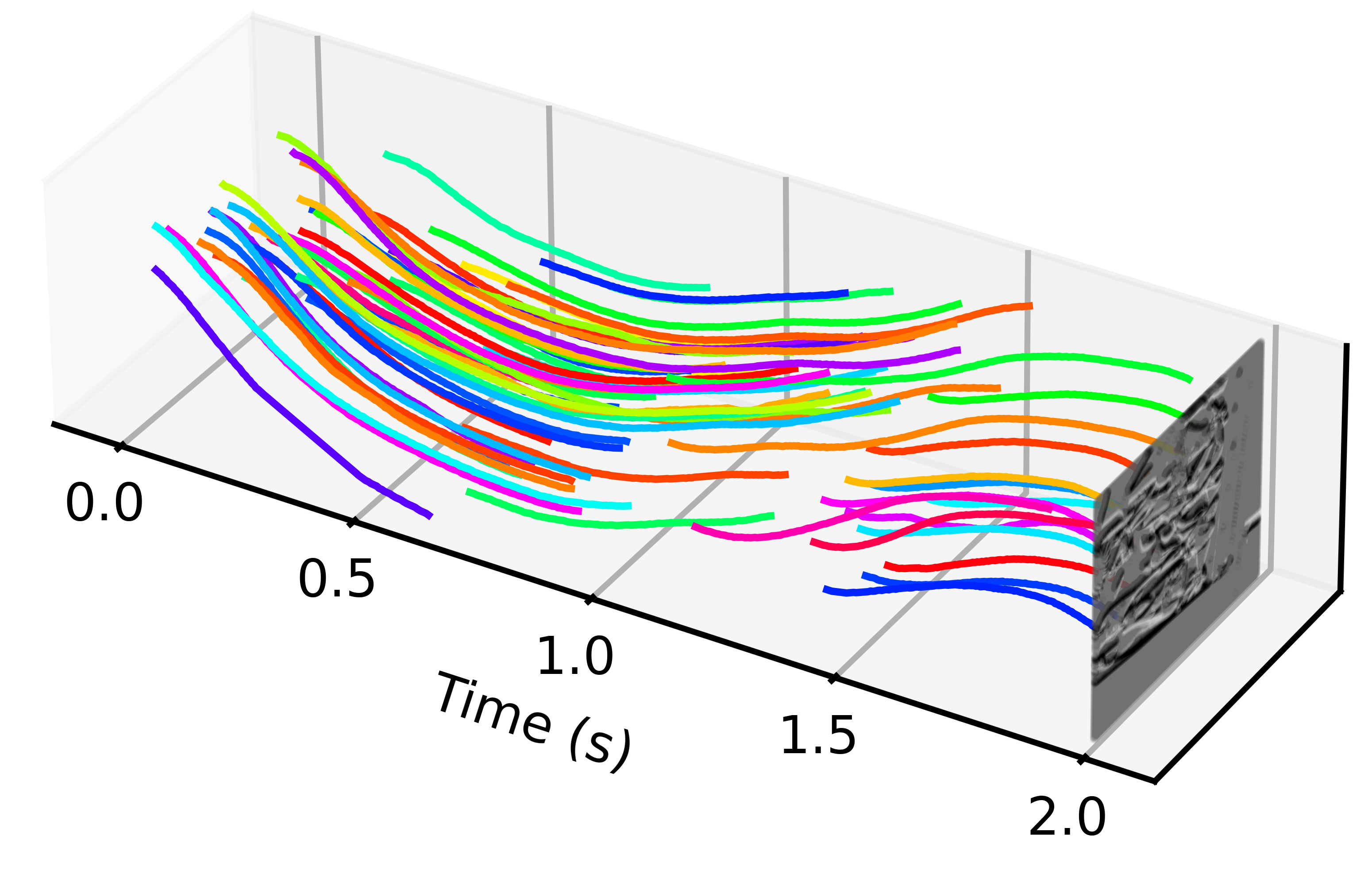}
    \\
    \includegraphics[width=.19\linewidth]{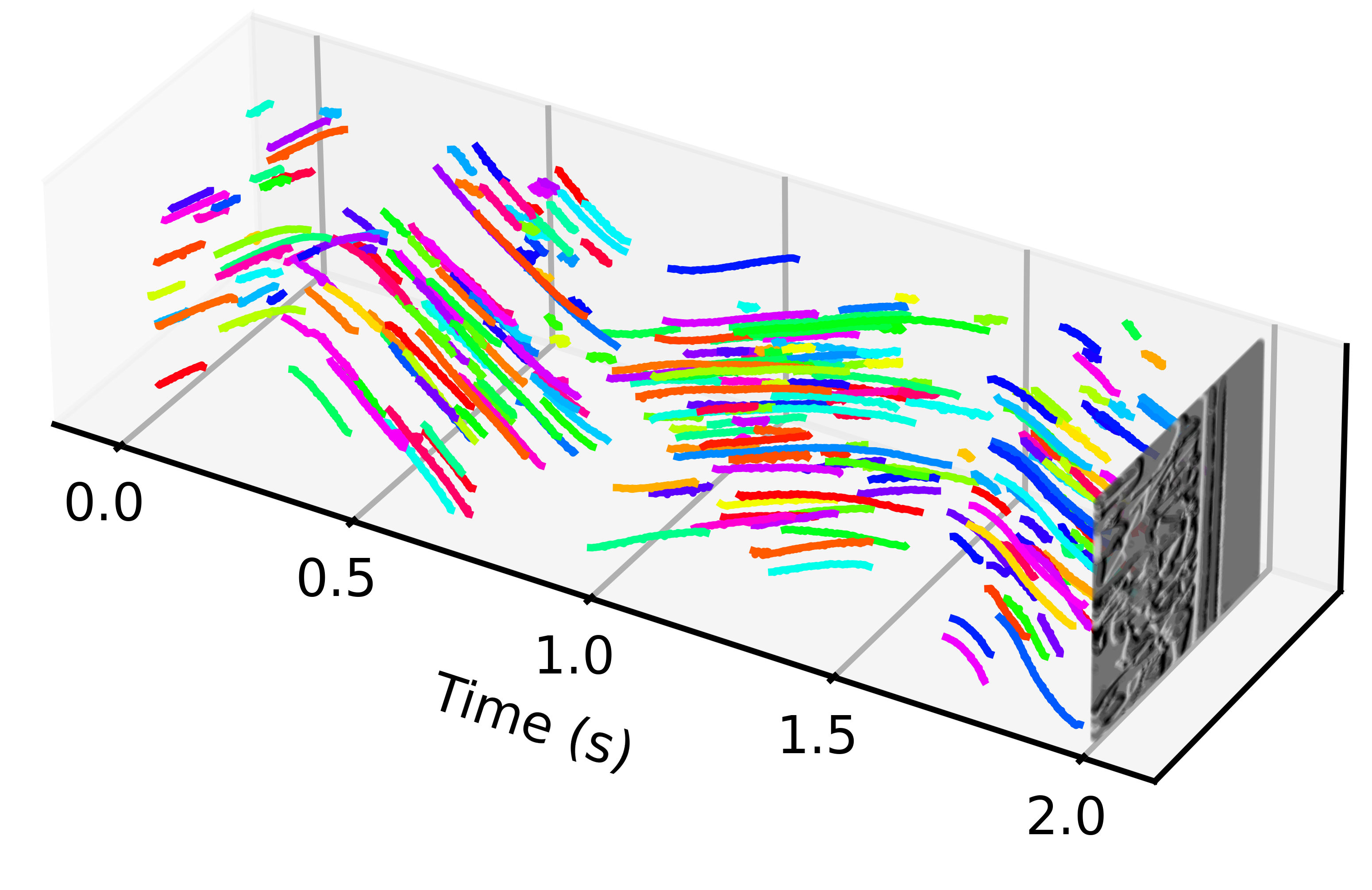}&
    \includegraphics[width=.19\linewidth]{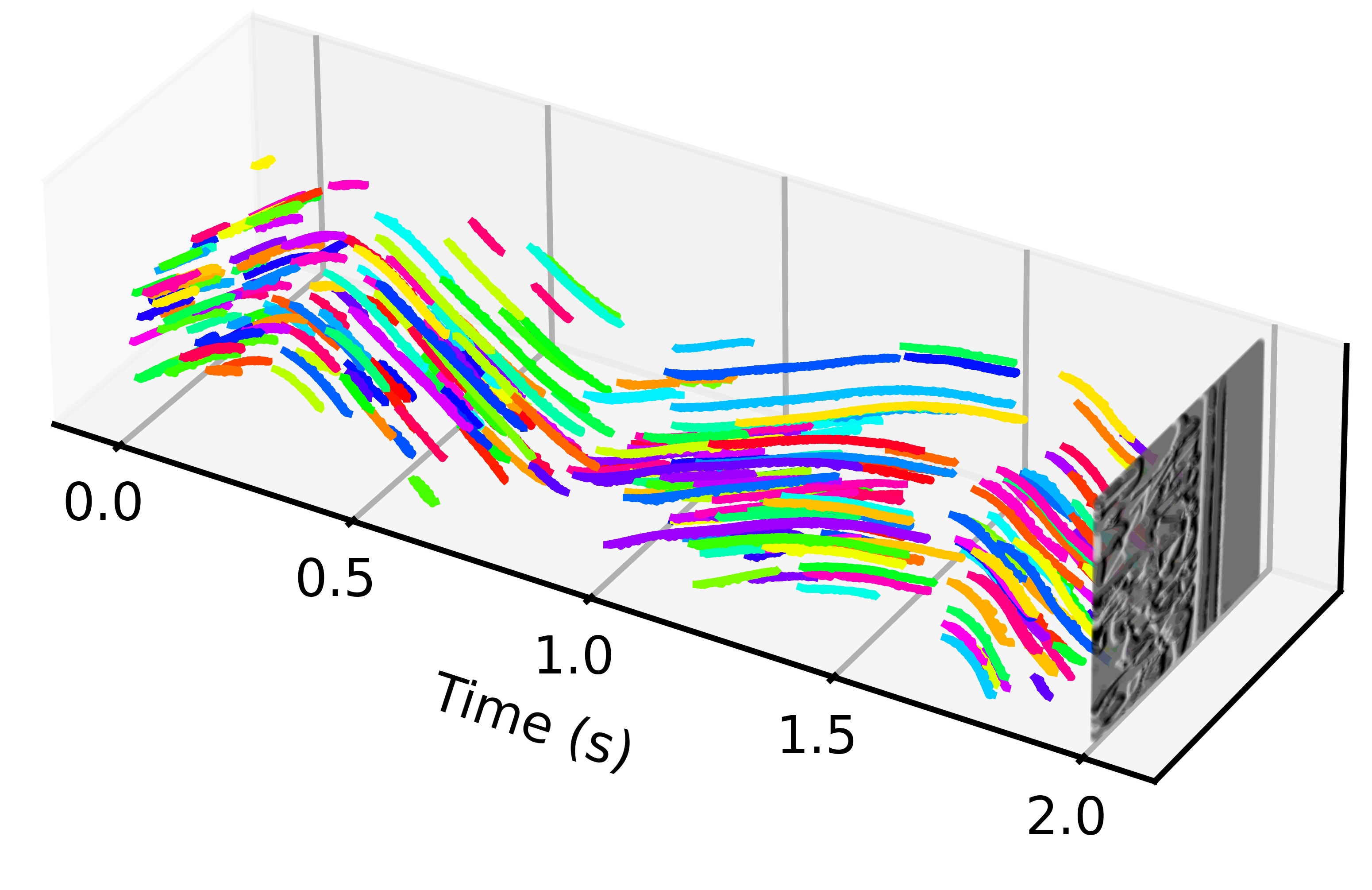}&
    \includegraphics[width=.19\linewidth]{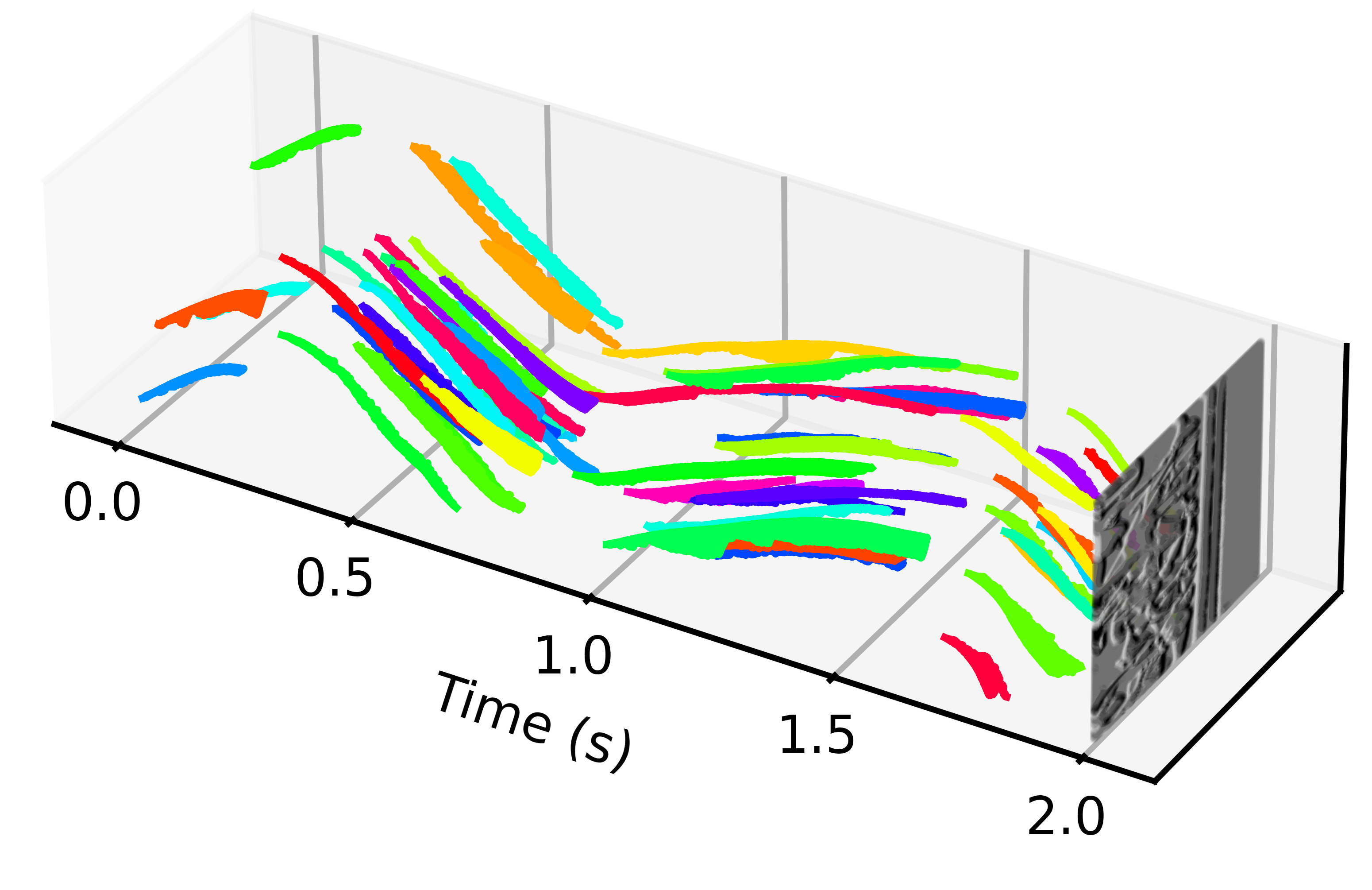}&
    \includegraphics[width=.19\linewidth]{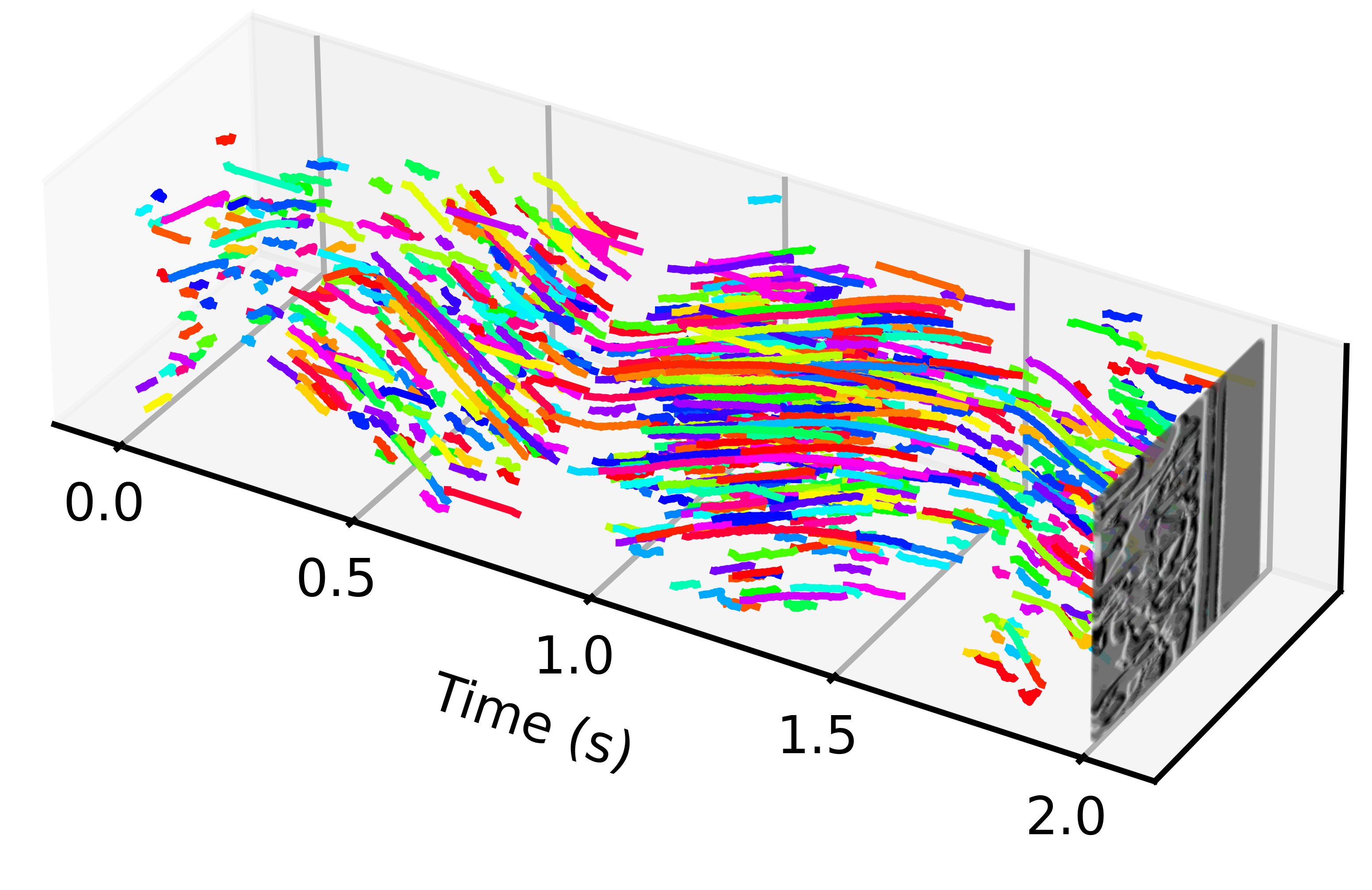}&
    \includegraphics[width=.19\linewidth]{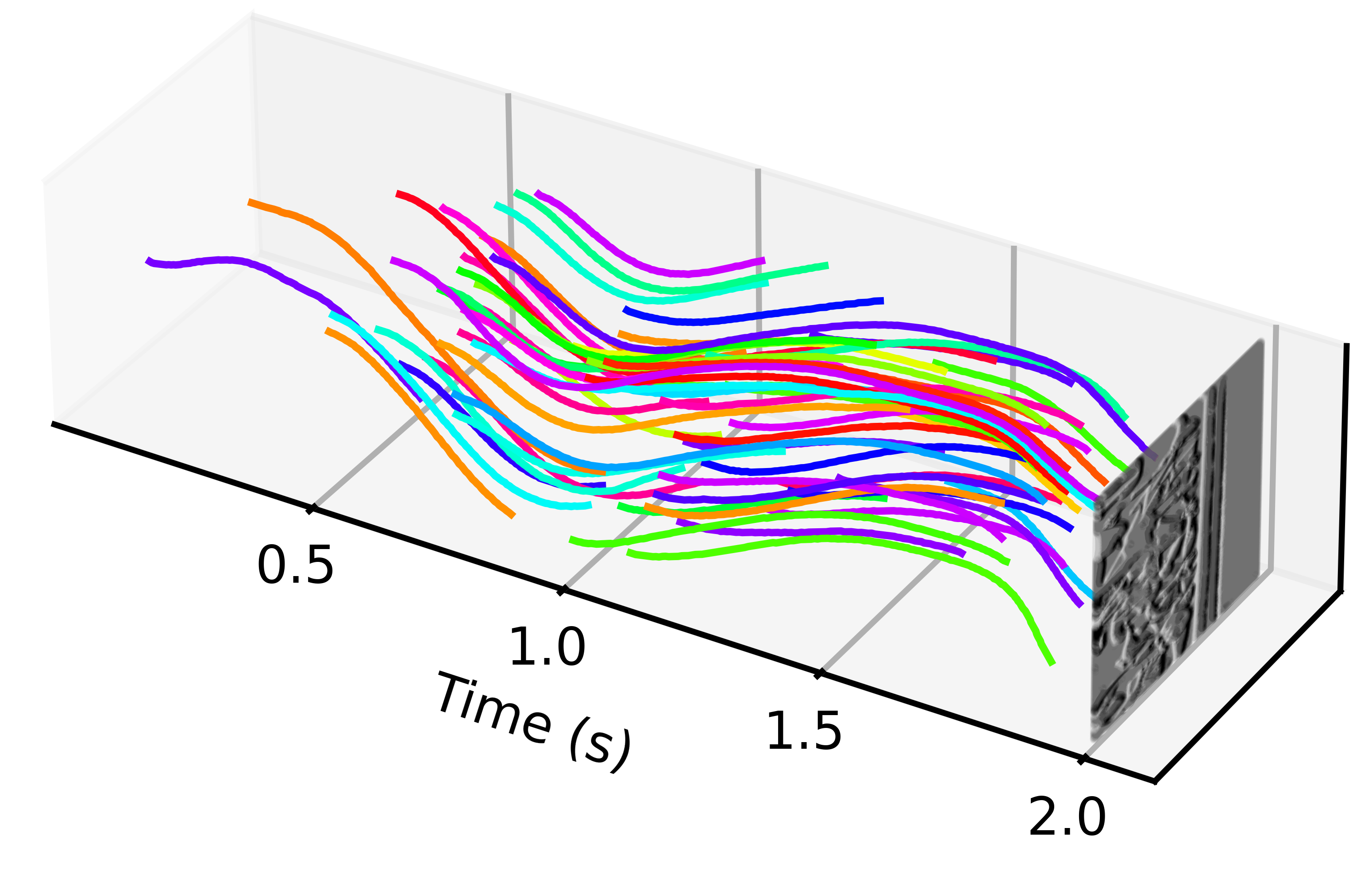}
    \\
    \includegraphics[width=.19\linewidth]{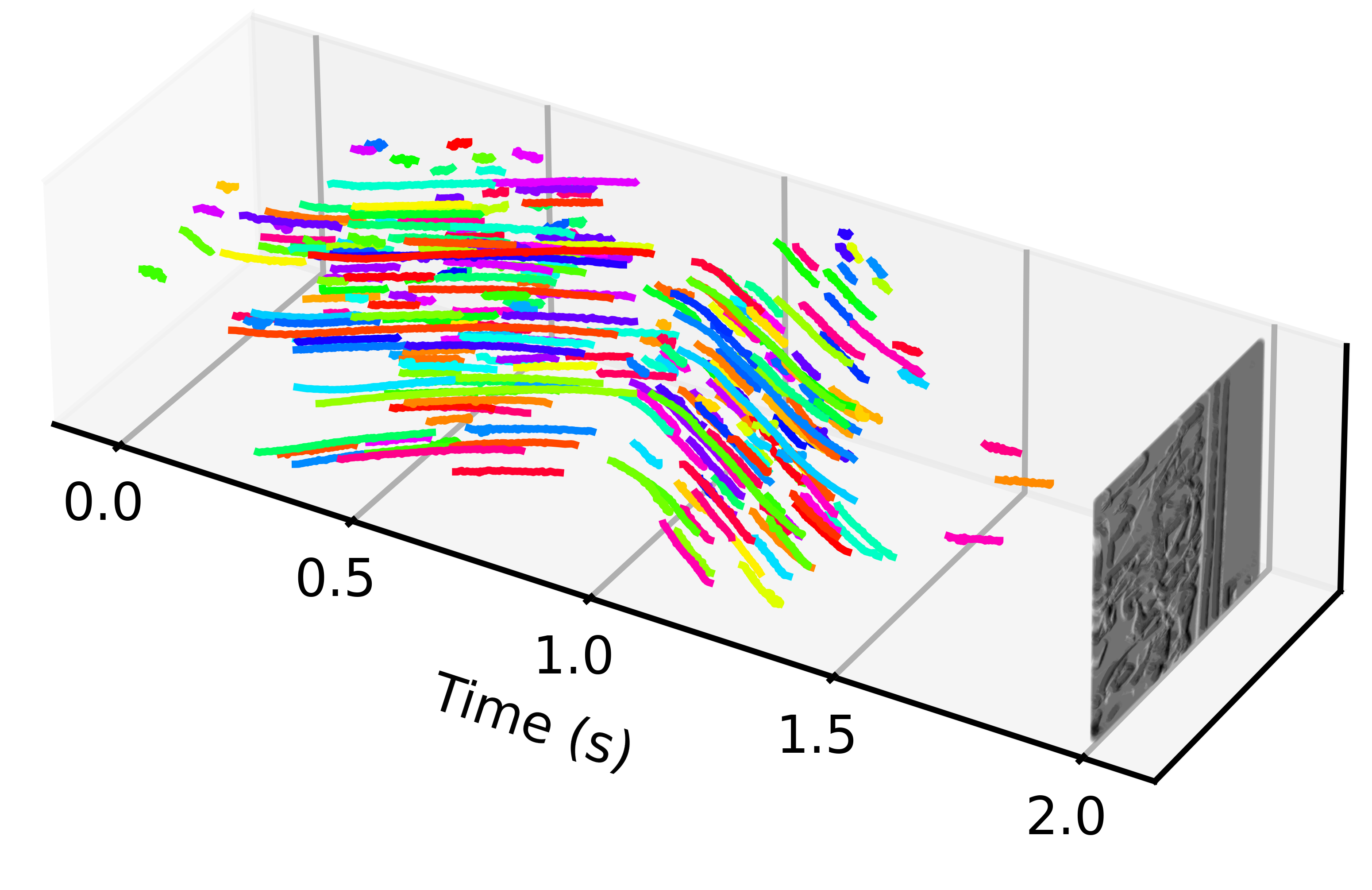}&
    \includegraphics[width=.19\linewidth]{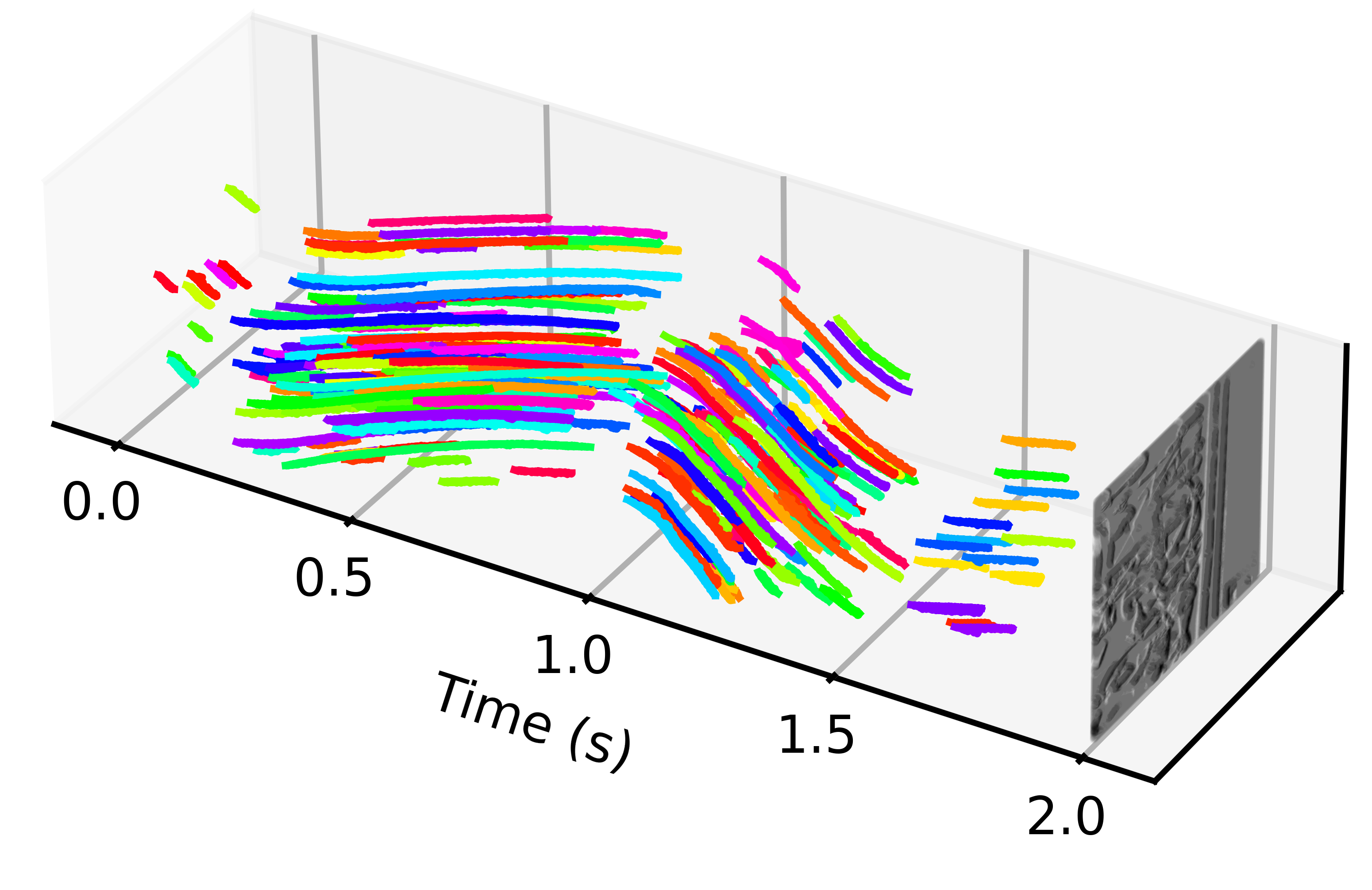}&
    \includegraphics[width=.19\linewidth]{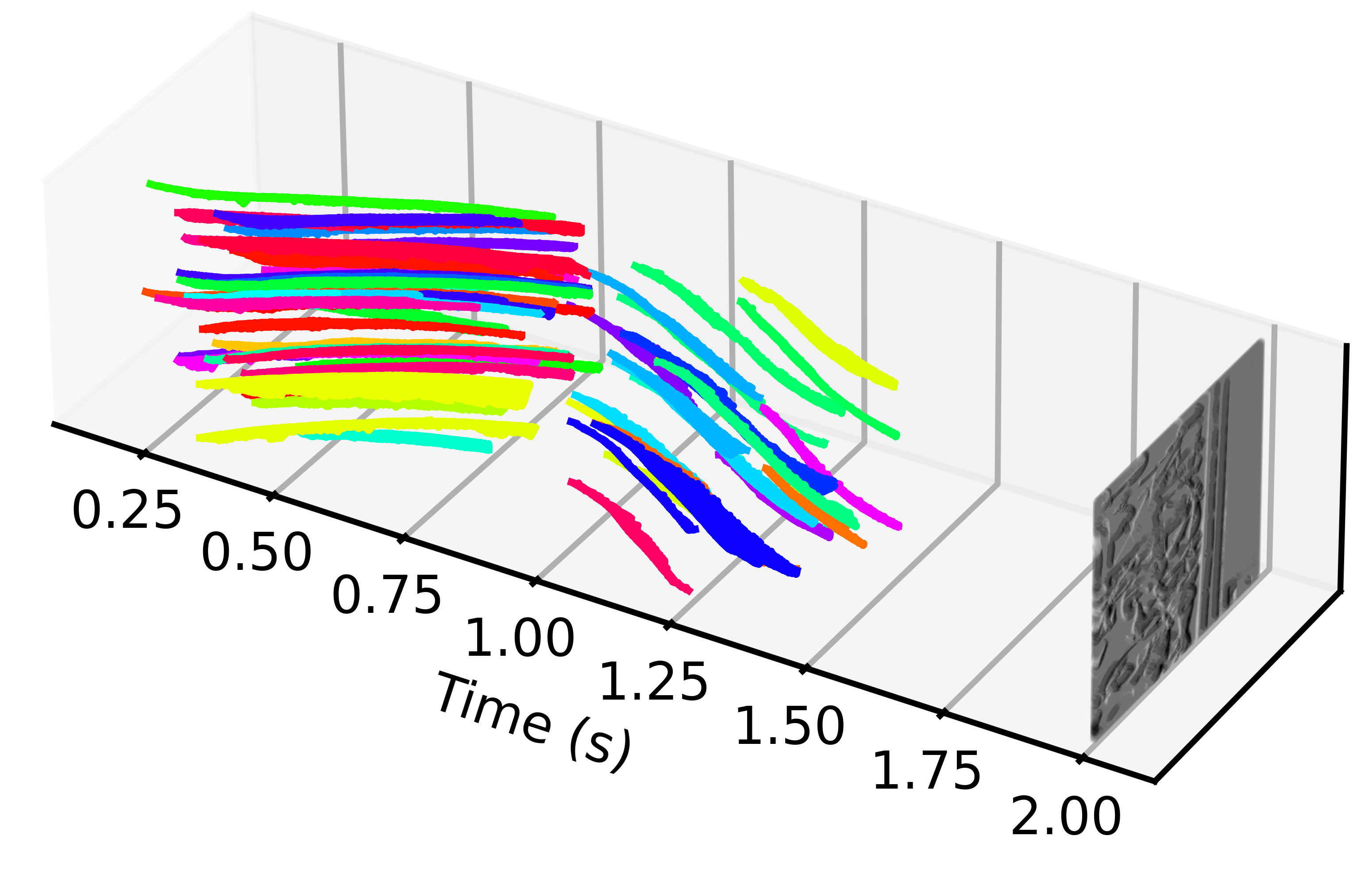}&
    \includegraphics[width=.19\linewidth]{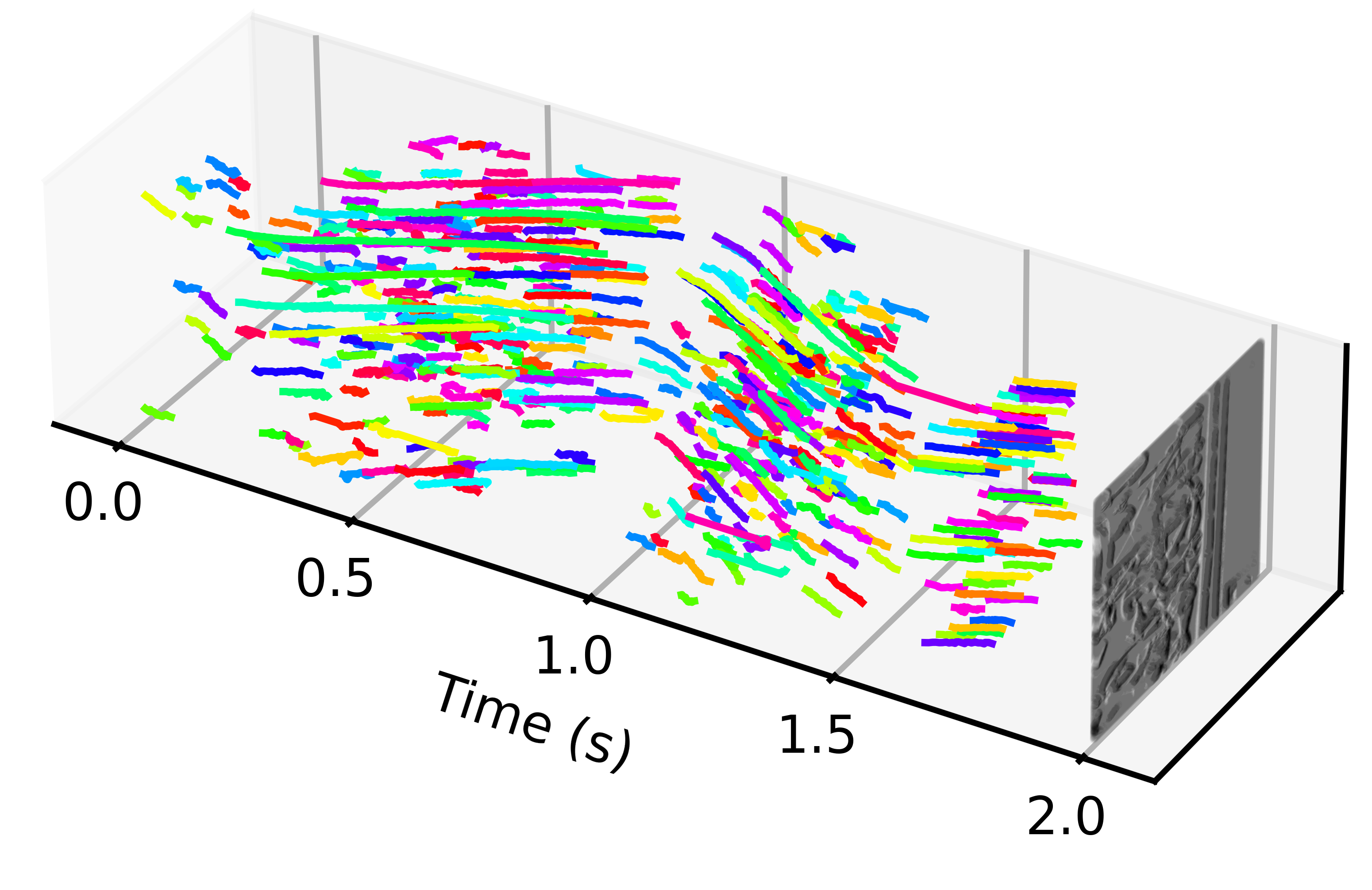}&
    \includegraphics[width=.19\linewidth]{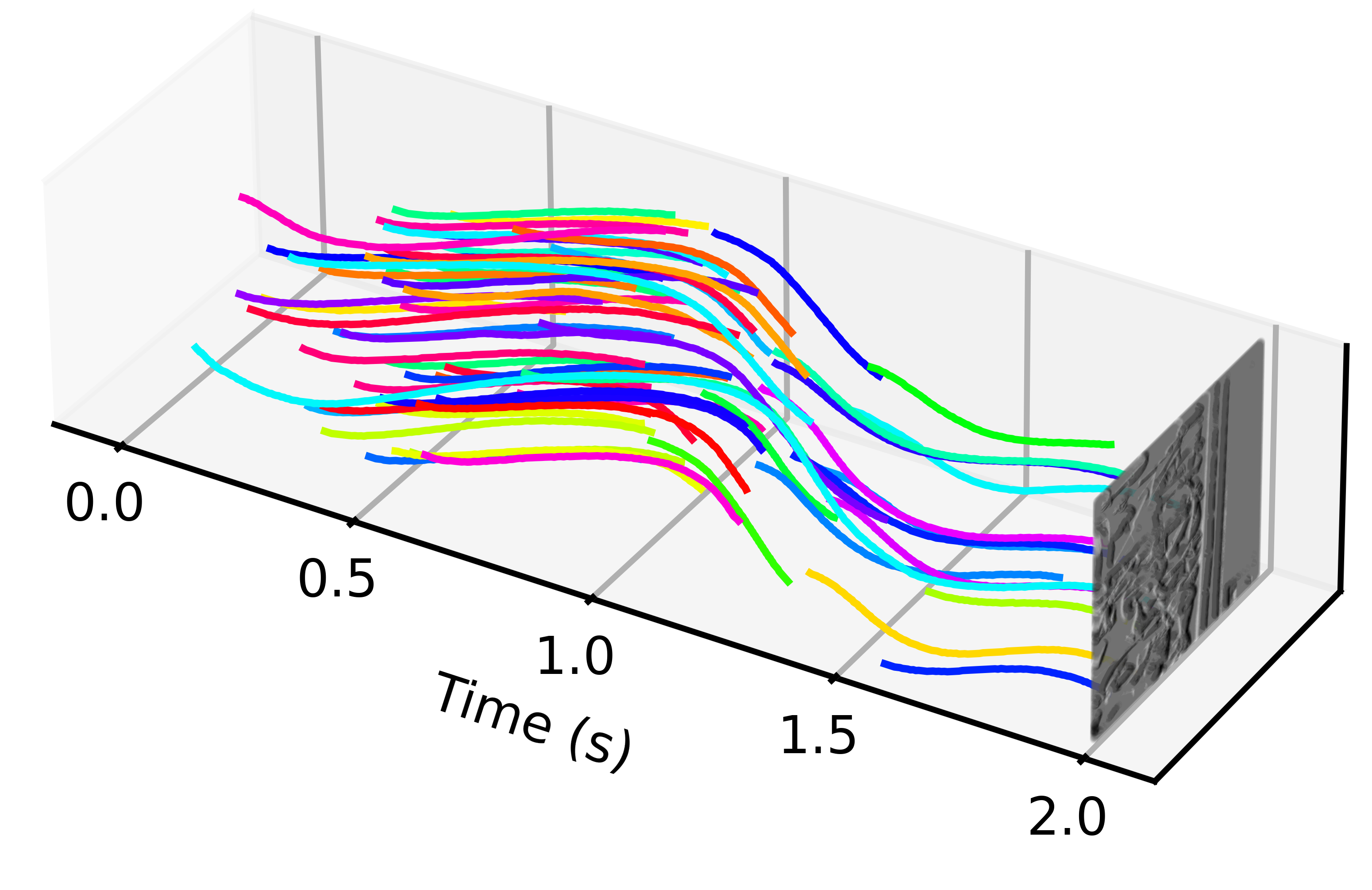}
    \\
    eHarris &  eFast & Arc & luvHarris & Ours \\
\end{tabular}    
    \caption{\label{fig:qualitative_results} {\bf Visual comparisons
        of tracks obtained by different methods on the HVGA ATIS
        Corner Dataset. }
      The tracks obtained with our detector are significantly longer.
}
\end{figure*}

Table~\ref{table:comparison_others} reports the comparison results on the HVGA ATIS Corner Dataset. Our detector is able to beat previous methods both in terms of accuracy as given by the $\delta t$-homography reprojection error and of track lifetime metrics. Moreover, Figure~\ref{fig:qualitative_results} provides visual comparisons of the track lengths for these methods.

Table~\ref{table:zurich_results} reports the comparison results on The Event-Camera Dataset. While, as discussed at the beginning of this section, the results can be considered as biased, they show that our detector performs well on this other dataset captured with a different sensor.

\begin{table}[htbp]
\begin{center}
\begin{tabular}{c@{$\quad$}c@{$\quad$}c}
 \toprule
  Method & $P_\text{rel}$ & $R_\text{rel}$\\
 \hline
 eHarris \cite{eharris} & 1 & 1 \\
 eFast \cite{efast} & 0.68 & 0.55 \\
 Arc \cite{arc} & 0.84 & 0.59 \\
 luvHarris \cite{luvharris} & 0.86 & 0.97 \\
 ours & \textbf{1.03} & \textbf{2.17} \\
 \bottomrule
\end{tabular}
\end{center}
\caption{{\bf Evaluation on  The Event-Camera Dataset}. The dataset was captured with a different sensor than the HVGA ATIS Corner dataset, nevertheless our detector also outperforms earlier work. Following the protocol of~\cite{luvharris}, we report $P_\text{rel}$ and $R_\text{rel}$, the precision and recall relative to eHarris. \label{table:zurich_results}}
\end{table}

\subsection{Qualitative results}

Figure \ref{fig:corners_on_synthetic} shows how our network is able to predict heatmaps through time which are coherent with one another. The keypoint is correctly predicted at multiple locations for the same volumetric input despite long accumulation time which make the edges become thicker. Keypoint are correctly detected for easy corners on the left as well as more textured keypoint. The consistency through time is a great advantage of our method and make it possible to track corners using a simple nearest neighbor algorithm.
\\
Similarly, Figure \ref{fig:qualitative_results} demonstrate the homogeneous spatial distribution of our keypoints which create very long and smooth tracks. Our method helps with tracks continuity, they are seldomly lost or broken which is not the case for other methods such as eHarris, eFast, Arc or even luvHarris.   

\subsection{Computation Times and Memory Footprint}
Our network has less than 26K parameters and runs in 7ms on a GTX 1080 GPU, for a HVGA input event sensor. The low footprint of the network coupled with the computation of the event cube enable running at arbitrarily high event rate. Our detector offers the possibility to reduce even more computations by augmenting the integration period and the number of predicted heatmaps at the cost of only minimal latency. This makes our novel approach very suitable for real-time applications.

\section{Conclusion}

We presented a novel keypoint detector in event streams, which provides long tracks of accurately localized keypoints. We will make our code available, and believe it will be useful for many downstream applications, such as SLAM and object tracking.

We also believe that predicting as we do multiple successive estimates rather than one for the integration period used by many methods is general and could be applied to many other computer vision problems on event-based cameras, such as segment detection,depth prediction and object detection.

%
%
\clearpage 

{\small
\bibliographystyle{splncs04}
\bibliography{cleaned_egbib, vincents_refs}
}
\end{document}